\documentclass[5p, twocolumn]{elsarticle}

\usepackage{lineno,hyperref}
\modulolinenumbers[5]

\usepackage{amsmath,amssymb,graphicx}
\usepackage{subcaption} 
\usepackage{color}
\usepackage{tikz,pgfplots} 
\usetikzlibrary{calc}
\usepackage{booktabs}
\usepackage{array}
\usepackage{ulem}
\usepackage{fancyhdr}
\graphicspath{ {./ICIP/} } 
\newcolumntype{L}[1]{>{\raggedright\let\newline\\\arraybackslash\hspace{0pt}}m{#1}}
\newcolumntype{C}[1]{>{\centering\let\newline\\\arraybackslash\hspace{0pt}}m{#1}}
\newcolumntype{R}[1]{>{\raggedleft\let\newline\\\arraybackslash\hspace{0pt}}m{#1}}

\journal{Journal of \LaTeX\ Templates}

\begin{document}
\fancyfoot{}
\fancypagestyle{pprintTitle}{%
\lhead{Published in the Neurocomputing Journal}
\renewcommand{\headrulewidth}{0.0pt}
}
\begin{frontmatter}
\title{Diverse Single Image Generation with Controllable Global Structure\tnoteref{mytitlenote}}

\author{Sutharsan Mahendren$^*$, Chamira U. S. Edussooriya$^{*,\#}$, Ranga Rodrigo$^*$}
\address{$^*$Department of Electronic and Telecommunication Engineering, University of Moratuwa, Sri Lanka \\
$^{\#}$Department of Electrical and Computer Engineering, Florida International University, Miami, FL, USA}






\begin{abstract}
Image generation from a single image using generative adversarial networks is quite interesting due to the realism of generated images. However, recent approaches need improvement for such realistic and diverse image generation, when the global context of the image is important such as in face, animal, and architectural image generation.
This is mainly due to the use of fewer convolutional layers for capturing the patch statistics and, thereby, not being able to capture global statistics well. The challenge, then, is to preserve the global structure, while retaining the diversity and quality of image generation. We solve this problem by using attention blocks at selected scales and feeding a random Gaussian blurred image to the discriminator for training. We use adversarial feedback to make the quality of the generation better. Our results are visually better than the state-of-the-art, particularly, in generating images that require global context. The diversity of our image generation, measured using the average standard deviation of pixels, is also better.
\end{abstract}

\begin{keyword}
Single image generation, generative adversarial networks, scale-wise attention, adversarial feedback.
\end{keyword}

\end{frontmatter}




\section{Introduction}
\label{sec:intro}

Generative Adversarial Networks (GANs) are successful in implicitly learning the underlying statistics of a large dataset and thus enable generating new samples from the same distribution~\cite{GOODFE14},~\cite{radford2016unsupervised}. In such GANs, generating good-quality and diverse images needs a large image dataset. Recently, SinGAN~\cite{SHAHAM19} proposed a hierarchical learning-based approach  for training with a single image. Here, in each scale, the generator and discriminator with low receptive fields learn to capture the internal statistics of the patch distribution of the image. One of the drawbacks of this  method, as the  paper itself states, is unrealistic image results when the global structure of the image is important, e.g., in face and animal image generation.
The main reason for unrealistic results is the lack of global structure, when the images are generated starting from the coarsest scale. SinGAN can only generate the images without destroying the global structure by feeding the downsampled version of the real image in less-coarser scales instead of feeding noise to the coarsest scale. However, this tends to reduce the diversity in generation. 

The level of information that must be captured for the global structure of the image to produce realistic looking results varies with the input image. For example, the generated samples from the images with small structures in the foreground (e.g., balloons in air, flocks of birds) and natural scenes (e.g., landscapes, foliage) do not need to maintain its original global structure. In contrast, the global structure plays a major role in realistic generation of images of large objects like faces and buildings. 

\textcolor{black}{Prior works~\cite{SHAHAM19,hinz2021improved} in single image GAN domain depend on the receptive field of convolutional layers to incorporate local information to the network. Increasing the capacity of the network causes overfitting to the single sample and looses the diversity. Our main motivation is increasing the generation quality without degrading the diversity } 

In this paper, we propose two main strategies for inserting the global context to the network while maintaining the diversity of the image generation. First is using self-attention (SA) as a key to control the level of the insertion of information on the global structure for realistic generations of all type of images while not sacrificing much in the diversity. Second is using a random Gaussian kernel to convolve the real image before feeding it to the discriminator. This helps to improve the diversity in the generated images. Using these strategies we are able to generate diverse set of images starting from the coarsest scale with global context. \textcolor{black}{The} level of diversity in our results is significantly higher than the SinGAN, when we start the generation from less-coarse scales. In addition, to generate better reconstructions (in terms of Single Image Fr{\'e}chet Inception Distance (SIFID)) we leverage on the concept of adversarial feedback in our single-image multi-scale network. \textcolor{black}{Realistic images generated through our method are useful in tasks such as data augmentation~\cite{arxiv.1810.10863,rs13091713} and animation. Moreover, our method does not affect downstream tasks such as image editing, harmonization, and arbitrary size generation from the SinGAN architecture.}

The main contributions of this work are: 
(1) proposing a method that retains the global structure in generated images by using SA, 
(2) increasing the diversity of generated images by depriving the discriminator of high frequency detail by simple random Gaussian smoothing. 
(3) enabling the controllability of the global structure by carefully choosing where SA is used and by the standard deviation of the Gaussian kernel. (4) Using adversarial feedback from previous scales to generate better reconstructions in a single-image multi-scale network. To the best of our knowledge, this is the \textit{first time}, feedback from the discriminator is utilized under a multi-scale architecture.

\subsection{Related Work}


The ability of a GAN to generate a sample from a distribution resembling data resulted in many contributions in GAN-based image generation~\cite{karras2019style}, \cite{radford2016unsupervised}. These GANs generate novel images by learning from a large database of images, that would have emanated from the same distribution. StackGAN~\cite{zhang2017stackgan} and ProgressiveGAN~\cite{karras2018progressive} use progressively growing architectures to improve the stability while generating high resolution images.  BiGGAN~\cite{brock2018large} trains with large number of parameters and a large batch size on ImageNet dataset to attain high fidelity generation. Although GANs generate impressive results, the need for a large dataset, the resulting large training time, dataset specific nature of the generation are concerns. 

As our objective in this work is single-image GAN generation, here we concentrate on {\bf Deep Internal Learning}: Training a deep architecture with a single image for image specific tasks comes under deep internal learning. Deep  Image Prior (DIP)~\cite{Ulyanov_2018_CVPR}  captures the image statistics through the structure of the generator network to perform image restoration tasks like denoising, inpainiting, and super-resolution. Here, the  network  learns to map random input to the single image sample. In this reconstruction process, DIP is able to recover the corrected version (denoised, inpainted) of the trained sample. Double DIP~\cite{DoubleDIP} extends this idea to decompose a single image into two with a task-specific mask and regularization. In segmentation, the image is decomposed into foreground and background with a binary mask. In image dehazing, the image is decomposed into an airlight map and haze free image with a transmission map. In ZSSR~\cite{ShocherCI18} an image specific CNN is learned to super resolve an image. It is trained with high and low resolution pairs from a single image. KernelGAN~\cite{Bell-KliglerSI19} proposes a method to extract a downsampling kernel from a deep linear convolutional generator and combine with ZSSR to super-resolve an image. ~\cite{ZuckermanNPBI20} extends the ZSSR concepts to video domain for temporal super resolution tasks. 
These works establish that it is possible to learn from patches in a single image for multiple tasks.

Now, there are prominent approaches for using deep-internal learning to generate images from a single image, making the generation process much faster to learn.
Spatial GAN uses~\cite{jetchev2017texture} a fully convolutional generator and patch discriminator starting from 2D noise to generate an arbitrary sized texture image from single image. In~\cite{zhou2018nonstationary}, the generator generates larger images condition on small patches from the training image. Here, VGG~\cite{simonyan2015deep} based perceptual loss is used alongside with adversarial training. \cite{bergmann2017learning} uses structural noise at three different levels at local, global, and periodic part at the generator with a patch discriminator. InGAN~\cite{Shocher_2019_ICCV} learns to remap an image to different aspect ratios and sizes while maintaining the same patch distribution by training a GAN based architecture with a multi-scale patch discriminator and a generator with a non-parametric transformation layer which can also learn to remap the output to the input using an inverse transformation. These networks typically map images to images and are, therefore, constrained to a couple of tasks. SinGAN, on the other hand, extends this deep internal learning concept with a hierarchical architecture to train from single image for multiple purposes. Since SinGAN only trains with a single image, it only learns statistics of the patches under its lower receptive field through a few convolution layers. Having a deep architecture with a higher receptive field will easily memorize the trained image and generate less diverse outcomes. In our work, we aim to improve SinGAN by having a controllable global structure insertion while maintaining the diversity of generation.

Although regular GANs, generate diversified images due to the diversity of the training set, single-image GANs suffer from the lack of diversity.
In prior work related to GANs on large training samples, loss of diversity in generation is considered as mode collapsing. \cite{Salimans2016} proposes mini batch discrimination to overcome mode collapse in training, where the similarity between intermediate features from discriminator for real and generated samples is used as additional information to discriminate the real from fake.  However, in the single-image generation context, there is no way to incorporate this similarity. \cite{Yang2019} proposes regularization to maximize the generator gradient with respect to the latent noise, specifically for conditional generation. While this may be of use, we opt to add the diversity through Gaussian smoothing at coarser scale  without using an additional regularization loss. \cite{Gurumurthy2017} uses latent noise from  a mixture model with learnable means and sigmas to improve the diversity. This impose more complexity for other task than generation such as editing and harmonization.
 
Several prior works leverage different kinds of feedback---some with GANs---to improve the output results in deep learning. \cite{7780767,Carreira2016,9358229} use the output of the CNN as an input in an iterative manner to refine the results in different applications like instance segmentation, human pose estimation and medical image segmentation. \cite{DBPN2018} uses error feedback inside the feature space from back projection units from low to high and vice versa. \cite{Shama_2019_ICCV} and \cite{Huh_2019_CVPR} are first to explore the concept of feed back in GANs. They use discriminator feedback to improve the generation quality. \cite{Shama_2019_ICCV} uses the intermediate features of the discriminator to modify the corresponding features in the generator. First the network is trained without the feedback modules and then the generator is fixed while discriminator and feedback modules are allowed to be updated. \cite{Huh_2019_CVPR} uses adaptive spatial transform layer to use the generated results from previous iteration and its discriminator scores to find the affine parameter to modify the encoded channel features in the generator at the next iteration. Both of these methods have not been explored with multi scale architectures and in the context of single image GAN. Instead of using the discriminator feedback to improve the results in the same scale, we pass it to the generator in the next scale to improve the results in most challenging regions. 

Several works related to GANs show hierarchical training on multiple scales with different resolutions of images helping to achieve high quality, high resolution samples~\cite{NIPS2015_aa169b49,karras2018progressive,8237891}. The use of patch discriminators, where they infer each small patches inside the generated image as fake or real, has been explored in~\cite{demir2018patchbased,ISOLAP17}. 
\cite{8578843, 10.1007/978-3-030-01234-2_1,8100150} incorporate attention in the vision models by computing features in channel or spatial dimension using the global pooling operation. In our work we use self attention for incorporating global structure through finding the similarities of pixels at different locations on the trained image.
\cite{Zhang2019} proposes to use SA in both the generator and discriminator to capture the long range dependencies. The vision transformers apply global self-attention to full-sized images \cite{DOSOVI21} using $16\times 16$ patches. It uses multi-head self-attention blocks inside the transformer based architecture for classifying images. 

Recently proposed SinGAN~\cite{SHAHAM19} uses a hierarchical unconditional GAN approach with single image for performing many tasks like image generation, super-resolution, editing and harmonization. ConSinGAN \cite{hinz2021improved} proposes to concurrently train several scales in SinGAN to increase the conformity of generated images. 
\textcolor{black}{PatchGenCN~\cite{zheng2021patchgencn} explicitly models the internal distribution of patch statistics with hierarchical energy based models using a patch convolutional network. Above works contain the generator with only a stack of a few convolution layers which does not provide the global information at coarser scales. To improve the realism of generated images from more complex structures, ExSinGAN~\cite{ExSinGAN} uses external information through GAN inversion at coarsest scale from pretrained BigGAN~\cite{brock2018large} and perceptual loss from pretrained VGG-19~\cite{simonyan2015deep}. MOGAN~\cite{MOGAN} needs an additional input to specify the region of interest for performing foreground and background generation separately. HP-VAE-GAN~\cite{NEURIPS2020_c2f32522} uses hierarchical patch variational autoencoder at coarser scales for diverse video and image generation. It also depends on the lower receptive fields of convolutional layers at coarser scales.}
\textcolor{black}{Above works} show limited performance in generating realistic images for which the global structure is important with higher generation quality \textcolor{black}{without using external information, }e.g., faces and buildings. External information mostly causes lesser visual diversity in the generation. The main reason for this limitation is that these networks do not have \textcolor{black}{explicit} controllable parameters to capture global information while training. Therefore, generating diverse images from single image that need better representation of the global structure still needs exploration.

\section{Method}
\label{sec:pagestyle}
\begin{figure}[t!]
    \centering
    \includegraphics[bb=150 300 500  700, scale=0.8]{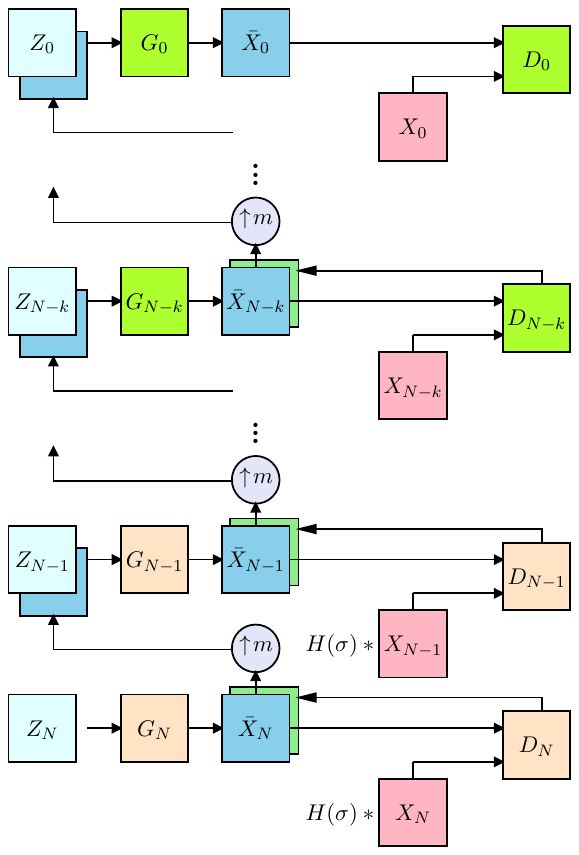}    
    \caption{Overall structure of our single image generator. See sec.~\ref{ss:brief_backgound} for details on the general single-image generation mechanism. Here, $G_0,D_0$ to $G_{N-k},D_{N-k}$ are convolutional blocks \textit{without SA}, and  $G_{N-k+1},D_{N-k+1}$ to $G_N,D_N$ are convolutional blocks \textit{with SA}. Further, $X_i$, $\bar{X}_i$ and $Z_i$ ($i=0,1,\ldots,N$) are real image, generated image, and $2$-D noise at scale $i$, respectively, and $\uparrow \!m$ denotes upsampling. $H(\sigma)$ is a Gaussian kernel. Notice that the scales with attention blocks  receive Gaussian smooth image as real samples for the discriminator. These infuse global context to image generation.}
    \label{fi:overall_architecture}    
    \vspace{-0.3\baselineskip}    
\end{figure}

We describe the proposed system in detail in this section. Fig.~\ref{fi:overall_architecture} shows the overall architecture of our system, which generates diverse images based on the training on a \textit{single image} while preserving the global structure. Our architecture contains several scales to train from coarsest to finer scales. In each scale, other than the coarsest one, the generator has the upsampled images generated from the previously trained scale, upsampled feedback from discriminator of previous scale, and noise as inputs. Here, each scale produces residual terms compared to the upsampled image from the previous scale. The generator at the coarser scale receive only 2D noise as an input. Our modification to the network architecture compared to SinGAN and \textcolor{black}{ConSinGAN} architectures is the introduction of SA blocks at the coarser scales in both the generator and discriminator and the use of adversarial feedback in all the scales above the coarsest one. Because we have a hierarchical structure and a patch-based discriminator, we can generate images based on a single input image. Furthermore, we can preserve the global structure of the input image, essential for intelligibility, due to the SA block. Feeding the input image (of appropriate scale) convolved with a Gaussian to the discriminator enables  us to achieve a good diversity in generated images. In summary, our single-image synthesizer is an unconditional GAN with a multiple-scale pipeline similar to SinGAN with a modified SA layers for the first few scales in both the discriminator ($D$) and generator ($G$). We use SA blocks to infuse global context into the image generation at selected scales.

To compensate for the reduction of diversity when we impose the long-range dependencies through SA blocks, we introduce a strategy to use Gaussian kernels with varying standard deviation (std.) to convolve the real image before feeding to $D$ at coarser scales only. With this, we can achieve the desired diversity while maintaining the global structure, which remedies one of the major issues in the SinGAN-like architectures.

\begin{figure}[t!]
    \centering
    \includegraphics[bb=150 600 500  720, scale=0.8]{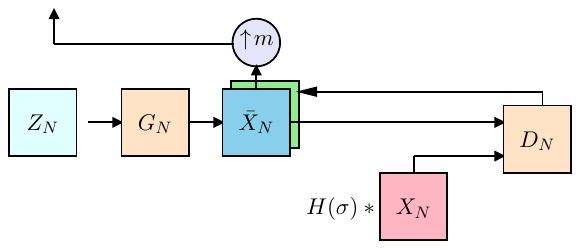}    
    \caption{Adversarial feedback: \textcolor{black}{the output from the discriminator of particular scale (e.g., $D_N$) is fed back through the upsampling block $\uparrow \!m$ to the generator $G_{N-1}$ of the next scale. }}
    \label{fi:adversarial_feedback}    
    \vspace{-0.3\baselineskip}    
\end{figure}

Fig.~ \ref{fi:adversarial_feedback} (an excerpt of Fig.~\ref{fi:overall_architecture}) shows how the adversarial feedback is used. \textcolor{black}{The output from the discriminator of} a particular scale (e.g., $D_N$) is fed back through the upsampling block $\uparrow \!m$ to the generator $G_{N-1}$ of the next scale. In this fashion, we are able to achieve adversarial feedback in our multi-scale architecture. Here, we add noise only for the upsampled image from the previous scale. In summary, with the inclusion of attention blocks at selected scales and using Gaussian convolution in the discriminator image inputs, we are able to produce a diverse set of images, particularly valuable when global structure is important.

\subsection{A Brief Background on Single Image Generation}
\label{ss:brief_backgound}
To create better understanding, here we briefly discuss about single-image generations. For more information on single image generation, following are good reads: \cite{SHAHAM19}, \cite{hinz2021improved} . The key idea to generating a generally similar but different image by learning from a single image is to learn the distribution of the patches within the image.  In each scale above the coarsest scale, the generator produces the residual term conditioned on upsampled version of generated image from previous scale (see Fig.~\ref{fi:intermediate results}). When the residual from the current scale is added to the upsampled term from the previous term a larger and diverse image is generated.

\subsection{Image Generation at Each Scale}

The procedure of image generation is the same as SinGAN. Coarsest scale generation starts with random noise. The generated results of each scale is fed to the scale above after upsampling. These scales add noise ($Z_{n-k}$) to the upsampled generation and use them as a prior to generate the image from the current scale. We attach the upsampled adversarial feedback with it. We can also generate images starting from scales lesser than $N$. To do that we downsample the original image to the size of the corresponding scale and feed that with added noise. In this situation adversarial feedback is computed from the scale above with the downsampled original image.

\subsection{Self Attention (SA) Block} 
\begin{figure}[t!]
    \centering
    \includegraphics[bb=0 0 400 140, width=0.98\linewidth]{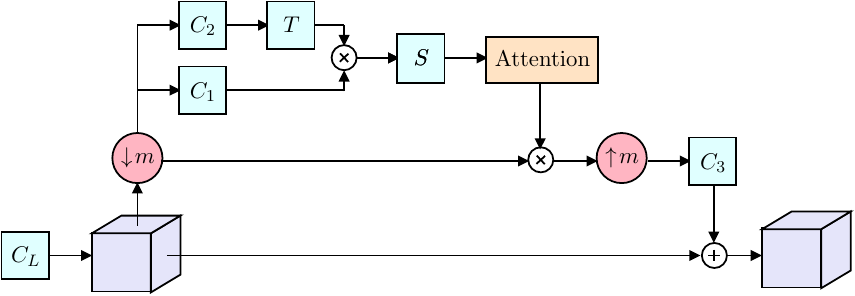}
    \caption{Self-attention block. This encourages the retention of the global structure in the generated image. Here, $C_L$ denotes the previous convolution layer, $C_1$, $C_2$ and $C_3$ are convolution layers. Further, $T$ denotes transpose and $S$ denotes softmax, and $\downarrow \!m$ and $\uparrow \!m$ denote downsampling and upsampling, respectively. This self attention block is added to the layers in generator and discriminator at coarser scales which are depicted at \textcolor{black}{ Fig.\ref{fi:overall_architecture} in orange tone.}}
    \label{fi:self_attention_block}    
    \vspace{-0.9\baselineskip}    
\end{figure}
The SA block, shown in Fig.~\ref{fi:self_attention_block}, is the important component that infuses global context to our image generation in addition to the convolution features which help to capture the information in the local neighborhood. The advantage of the SA block \textcolor{black}{
is that} it implicitly increases the receptive field of the features using the attention map which is computed considering all the features in space. It mitigates the need for a higher number of convolutional layers followed by a downsampling layers to increase the receptive field. 
\textcolor{black}{In hierarchical architectures, the coarser scales play the major role in capturing the global structure of the large objects inside the image. It is a much harder goal to achieve only by using stack of convolutional layers. Although adding self attention blocks helps to capture the global structure, it introduces small local artifacts in the generations. Due to this reason, we use the self-attention blocks in the coarser scales only. Rest of the scales in the GAN having only the convolution layers help to remove the noise and add more local details to the generation.  }

Unlike the implementation in ~\cite{Zhang2019}, we compute the SA features from the downsampled version of features from the last convolutional layer. In downsampling
map the features to the predefined fixed output size $m\times m$, $m \in [8, 16, 32$] irrespective of the input. Instead of using additional convolutions to compute the value features, we directly pass the downsampled features to be modified by the attention. Here, each feature is a weighted sum based upon its similarity with others in key and query features with an optional channel reduction. SA features are directly added to the convolutional feature without having a learnable parameter like in SAGAN~\cite{Zhang2019}. Experiments show this form is enough to capture the required global structure to pass it to the next scales.

\subsection{Increasing the Diversity with Gaussian Smoothing with Kernel with Random Standard Deviation}
\label{ss:gaussian_smoothing}
Adding SA to a greater number of scales capture more and more long-term dependencies inside a single image. However,  it reduces the diversity in the generation. Depending on the nature of the image, particularly the comparative size of the global structure (e.g., some faces, Eiffel tower image), we need to have SA in more scales than for others. 
Then, the final realistic outcomes become less diverse. It is hard to tune other parameters like the level of downsampling size or the positions of the SA block inside each scale to balance the trade-off between realistic generation and diversity. SinGAN uses the same downsampled version of the real image to be fed to $D$ depict the real sample for discriminator. It is one of the main reasons why SinGAN architecture has to maintain a lower receptive field with lesser layers to reduce the overfitting and loss in diversity. To solve this problem, we propose convolving with a Gaussian which can maintain both image quality and diversity: Instead of feeding the same image as the real sample, we feed it after convolving with a Gaussian kernel with random value for the std. sampled from a uniform distribution between a predefined min. and max. value. For the computation of the reconstruction loss, we keep a fixed std. value for generating the real sample.


\subsection{Adversarial Feedback}
As the discriminator too holds a significant amount of information on the distribution of patches, making the generator aware of the discriminator's spatial information improves the reconstructions quality~\cite{Shama_2019_ICCV}. 

To achieve this feedback information transfer, we concatenate the discriminator's score for each patch from the previous scale with the generator input for the current scale. 
\textcolor{black}{Generators at higher layers  (above the coarsest scale) only generates the residual images from the upsampled generation of the previous scale. Generation at the coarsest scale does not depend on adversarial feedback because it only depends on the noise that we feed. We connect the feedback to the scales above it. Diversity of the generations highly depends on the variations in the coarsest scale. Scales above the coarsest scale help to add more details on its upsampled versions by generating the residuals. In the higher scales we do not use self attention. So the concatenated feedback in higher scales only affect the results in the local neighbourhood because of the few convolutional layers. Therefore, in our approach, feedback does not degrade the diversity significantly while improving the generation quality.}


\subsection{Loss}

Eq.~\ref{eq:gan_loss} shows the original GAN loss used in Goodfellow \textit{et al.}~\cite{GOODFE14}. 
\begin{equation}\label{eq:gan_loss}
    \begin{split}
    \min_{G}\max_{D} \mathcal{L}_\text{adv}(G, D) &=  \mathbb{E}_{x \sim P_r} [\log(D(x)]\\ &+  \mathbb{E}_{z \sim P_z} [\log(1 - D(G(z)))] \\
    \end{split}
\end{equation}
Here, $D(x)$ is output of discriminator for real images which aims to give a probabilistic score for $x$ that belongs to the real data distribution, $G(z)$ is the fake image from the random vector $z$. Typically, several iterations of interchangeable generator and discriminator optimizing happen until they reach Nash equilibrium. This leads to the generator fully approximating the the distribution of data $P_r$ and discriminator being unable to differentiate between the two distributions. See  Goodfellow \textit{et al.}~\cite{GOODFE14} for more details. 
\begin{equation}\label{eq:infinimum}
    \begin{split}
    W(P_r,P_g) &=  \inf_{\gamma \in \pi(P_r,P_g)} \mathbb{E}_{(x,y) \in \gamma} [\|x - y\|] \\
    \end{split}
\end{equation}
\begin{equation}\label{eq:supremum}
    \begin{split}
    W(P_r,P_g) &=  \max_{w \in W} \mathbb{E}_{x \in P_r} [f_w (x)] - \mathbb{E}_{x \in P_g} [f_w (x)] \\
    \end{split}
\end{equation}
When the discriminator is optimal, the original GAN loss (Eq.~\ref{eq:gan_loss}) relates to the Jensen–Shannon (JS) divergence between the real and fake distributions. Since real and fake distributions mostly lie in a lower dimensional manifold, it may contains non overlapping regions. This scenario make the discriminator to learn separate real and fake images easily and gradients of JS divergence become very smaller for the generator. To overcome this issue WGAN~\cite{pmlr-v70-arjovsky17a} proposes Wasserstein distance instead of using JS divergence bases loss function.
Wasserstein distance is defined as the minimum of effort to move from one distribution to another among a possible joint distributions which have the marginals as real and fake distributions ($P_r, P_g$) are the options for each transport plan as in Eq.~\ref{eq:infinimum}. WGAN~\cite{pmlr-v70-arjovsky17a} authors use  Kantorovich-Rubinstein duality to evaluate the Wasserstein distance by formulating the discriminator as a parameterized family of functions ($f_w (x)$) with the k-Lipschitz constraint which maximizes the difference between the expected socres for real and fake images by updating the discriminator weights  as in Eq.~\ref{eq:supremum}. \cite{NIPS2017_892c3b1c} introduce gradient penalty loss term to impose the 1-Lipschitz constraint instead of using weight clipping in \cite{pmlr-v70-arjovsky17a}.

We use the WGAN-GP~\cite{pmlr-v70-arjovsky17a}~\cite{NIPS2017_892c3b1c} for the adversarial loss and reconstruction loss to make the network generate the real samples from a particular fixed sample of noise at each scale, modified to accommodate the convolution with the Gaussian $H(\cdot)$.
For each scale (see Fig.~\ref{fi:overall_architecture}) $n (< N)$
\begin{align*}
    L &= \min_{G_n}\max_{D_n} \mathcal{L}_\text{adv}(G_n, D_n) + \alpha\mathcal{L}_\text{rec}(G_n)\\
    \mathcal{L}_\text{adv} &=  \mathbb{E}_{\vec{x}_f \sim P_g} [D(\tilde{\vec{x}})] - \mathbb{E}_{\vec{x}_r \sim P_h(\sigma_1, \sigma_2)} [D(\tilde{\vec{x}})] \\
    &+\lambda \mathbb{E}_{\hat{\vec{x}} \sim P_{\hat{\vec{x}}}} \left[( \|\nabla_{\hat{\vec{x}}} D(\hat{\vec{x}})\|_2 - 1)^2\right]\\
    \mathcal{L}_\text{rec} &= \|G_n(0, ({\vec{x}}_{n+1}^\text{rec}\uparrow^{r})  - \tilde{\vec{x}}_n\ast H(\sigma_3) \|^2
\end{align*}
where $P_h$ is the distribution of images convolved with a Gaussian, $\vec{x} \ast H(\sigma)$ and $\sigma \sim \mathrm{U}(\sigma_1, \sigma_2) $ and $\sigma_3 \in  [\sigma_1, \sigma_2]$.The convolution with the Gaussian $H(.)$ is for increasing the diversity of generation (while maintain the global structure). Note that convolution with $H(\cdot)$. is not used for scales without the attention block. See Sec.~\ref{ss:gaussian_smoothing} for more details.

\subsection{Selection of Parameters} 
Several parameters select the insertion level of global information. (1) No. of scales with SA starting from the coarsest scale ($k$). With this, SA is on scales $N$ to  $N-k+1$. (2) Choices of layers in each $G$ scale to add SA. (3) Max. and min. value for std. for the Gaussian kernel $\sigma_1$ and $\sigma_2$. (4) Output size after downsampling  with a factor $m$ inside the SA blocks\textcolor{black}{.} Items (1) (2) and (3) are impactful. The default choice for (4) is  $16\time 16$ which worked well for all the images that we tested. (1) and (2) directly control the global structure in the image generation. 

\section{Experimental Results}
We carried out experiments to show 1.  how the attention blocks and Gaussian smoothing of the input to the discriminator generate high-quality diverse images, 2.  the effect of the hyper-parameters ($k$ and $\sigma$), 3. impact of using Gaussian smoothing only\textcolor{black}{,} 4. impact of using adversarial feedback only, 5. overall impact of using feedback, self-attention, and Gaussian smoothing together comparing to SinGAN and ConSinGAN 6. how our system can perform editing, harmonization, and arbitrary-sized generation. 

We keep the LR of the generator ($G$) and discriminator ($D$) at $0.0001$ and train for $6000$ epochs with updating $G$ and $D$ one time in each epoch. We use $10$ for $\alpha$ as a weight for reconstruction loss. In each epoch, $G$ and $D$ are updated with the loss on a single real and fake pair. In our experiments we use instance normalization~\cite{UlyanovVL16Ins} instead of using batch normalization~\cite{pmlr-v37-ioffe15} as in SinGAN. We keep this configuration as baseline for SinGAN and our approaches. We also analyze ConSinGAN with our choice of parameters as mentioned above.

 \subsection{Impact of Self Attention Blocks and Gaussian Smoothing in Single Image Generation}
 
\begin{figure}[t!]
    \centering
    \begin{subfigure}[t]{0.24\linewidth}
        \includegraphics[width=\linewidth]{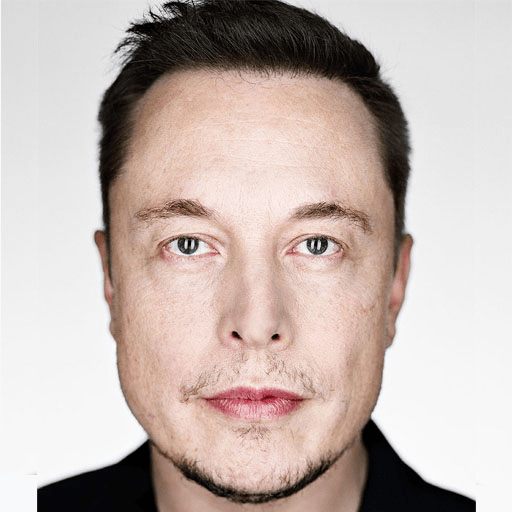}\\
        \vspace{-1.2em}
        \includegraphics[width=\linewidth]{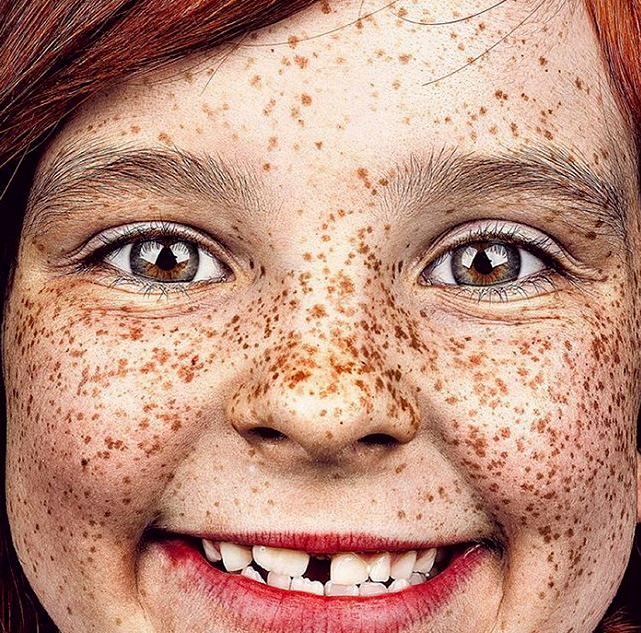}\\
        \vspace{-1.2em}
        \includegraphics[width=\linewidth, trim={0 10px 0 10px},clip]{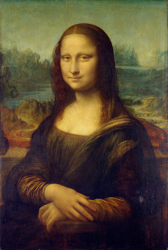}\\
        \vspace{-1.2em}
        \includegraphics[width=\linewidth]{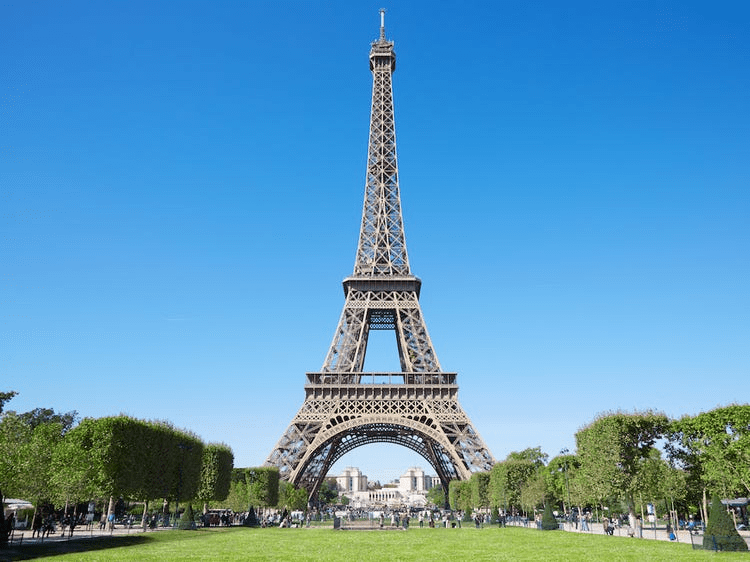}\\
        \vspace{-1.2em}        
        \includegraphics[width=\linewidth]{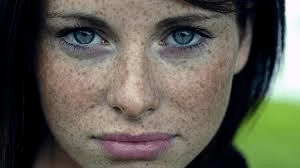}     
        \caption{Input}
    \end{subfigure}%
    \hspace{0.1em}%
    \begin{subfigure}[t]{0.24\linewidth}
        \includegraphics[width=\linewidth]{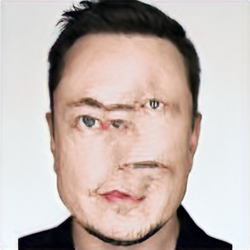}\\
        \vspace{-1.2em}
        \includegraphics[width=\linewidth]{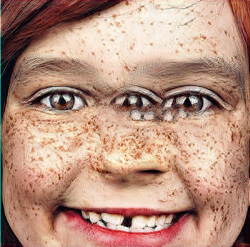}\\
        \vspace{-1.2em}
        \includegraphics[width=\linewidth, trim={0 10px 0 10px},clip]{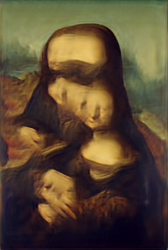}\\
        \vspace{-1.2em}
        \includegraphics[width=\linewidth]{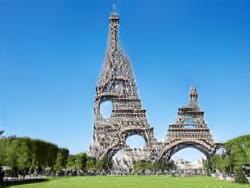}\\
        \vspace{-1.2em}
        \includegraphics[width=\linewidth]{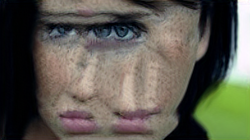}     
        \caption{SinGAN~\cite{SHAHAM19}}
    \end{subfigure}%
    \hspace{0.1em}%
    \begin{subfigure}[t]{0.24\linewidth}
        \includegraphics[width=\linewidth]{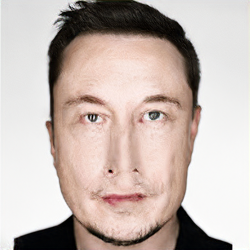}\\
        \vspace{-1.2em}
        \includegraphics[width=\linewidth]{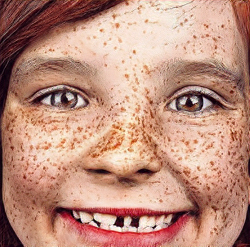}\\
        \vspace{-1.2em}
        \includegraphics[width=\linewidth, trim={0 10px 0 10px},clip]{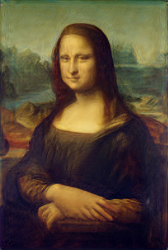}\\
        \vspace{-1.2em}
        \includegraphics[width=\linewidth]{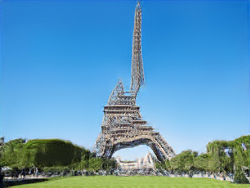}\\
        \vspace{-1.2em}
        \includegraphics[width=\linewidth]{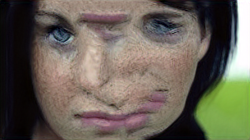}      
        \caption{$k=3, \sigma_c$}
    \end{subfigure}%
    \hspace{0.1em}%
    \begin{subfigure}[t]{0.24\linewidth}
        \includegraphics[width=\linewidth]{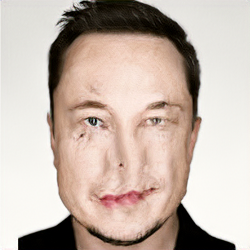}\\
        \vspace{-1.2em}
        \includegraphics[width=\linewidth]{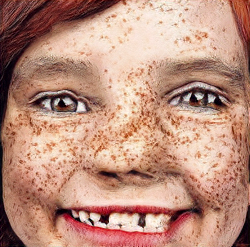}\\
        \vspace{-1.2em}
        \includegraphics[width=\linewidth, trim={0 10px 0 10px},clip]{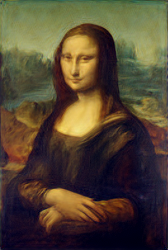}\\
        \vspace{-1.2em}
        \includegraphics[width=\linewidth]{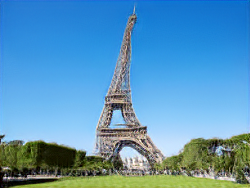}\\
        \vspace{-1.2em}
        \includegraphics[width=\linewidth]{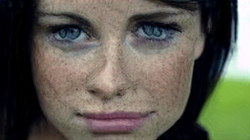}     
        \caption{$k=4, \sigma_d$}
    \end{subfigure} 
    \begin{subfigure}[t]{0.24\linewidth}
        \includegraphics[width=\linewidth]{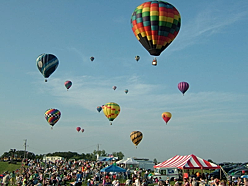}\\
        \vspace{-1.2em}
        \includegraphics[width=\linewidth, trim={0 18px 0 10px},clip]{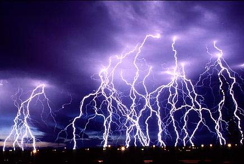}\\
        \vspace{-1.2em}
        \includegraphics[width=\linewidth, trim={0 10px 0 0},clip]{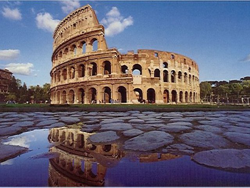}
        \caption{Input}
    \end{subfigure}%
    \hspace{0.1em}%
    \begin{subfigure}[t]{0.24\linewidth}
        \includegraphics[width=\linewidth]{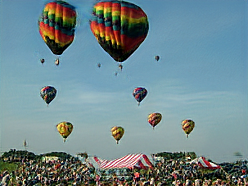}\\
        \vspace{-1.2em}
        \includegraphics[width=\linewidth, trim={0 10px 0 10px},clip]{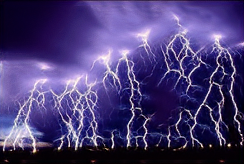}\\
        \vspace{-1.2em}
        \includegraphics[width=\linewidth, trim={0 10px 0 0},clip]{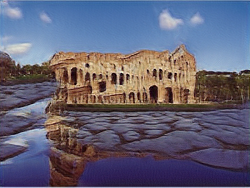}
        \caption{SinGAN~\cite{SHAHAM19}}
    \end{subfigure}%
    \hspace{0.1em}%
    \begin{subfigure}[t]{0.24\linewidth}
        \includegraphics[width=\linewidth]{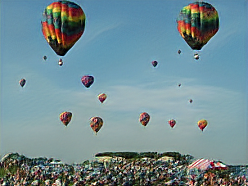}\\
        \vspace{-1.2em}
        \includegraphics[width=\linewidth, trim={0 10px 0 10px},clip]{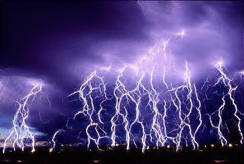}\\
        \vspace{-1.2em}
        \includegraphics[width=\linewidth, trim={0 10px 0 0}, clip]{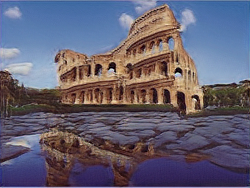}
        \caption{$k=1, \sigma_g$}
    \end{subfigure}%
    \hspace{0.1em}%
    \begin{subfigure}[t]{0.24\linewidth}
        \includegraphics[width=\linewidth]{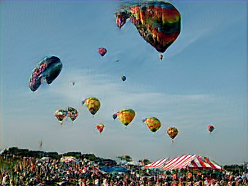}\\
        \vspace{-1.2em}
        \includegraphics[width=\linewidth, trim={0 10px 0 10px},clip]{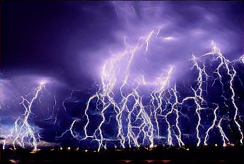}\\
        \vspace{-1.2em}
        \includegraphics[width=\linewidth, trim={0 15px 0 0},clip]{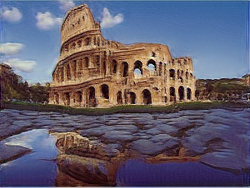}
        \caption{$k=3, \sigma_h$}
    \end{subfigure}%
    \caption{Here we compare the baseline with only using Gaussian smoothing and attention. Attention retains the global structure. Attentions needs to couple with Gaussian smoothing to have diversity in results. (a) to (d): for images that needs to maintain the global structure. (e) to (h): for images that do not need to maintain the global structure. (c), (d), (g) and (h): our results with two sets of hyper-parameters. $\sigma_c \in [1,3]$,  $\sigma_d,\sigma_g,\sigma_h \in [3,7]$.  Notice that our method retains the global structure better than SinGAN~\cite{SHAHAM19} for the images which require global structure for its realistic look.}
    \vspace{-0.9\baselineskip}
    \label{fi:resutls_global_and_without}                    
\end{figure}

 Here we  \textcolor{black}{qualitatively} and quantitatively explain how the self attention helps to increase the generated image quality alongside with Gaussian smoothing for increasing the diversity from single image.
Fig.~\ref{fi:resutls_global_and_without} compares our results with SinGAN.  Top five rows show images that need the global structure to be realistic. In column c and d we show our results with its hyper-parameters. We increase the number of scales with SA standard deviation values from column c to column d,  and add self-attention to first four layers inside G and D for the results in column d. Note that our results in column c is visually better than SinGAN for the image in the first three rows. For the images in the 4th and 5th row, our results in column d performs better as these images required another SA layer to recover the global structure and a higher $\sigma$ to maintain the diversity. Last three rows show images that do not need the global structure to be realistic. In column g where we use SA in the coarsest scale only. Even in this scenario our method produces diverse images on par with SinGAN. However, while we increase the number of scales with SA to 3 in column h, diversity becomes low as the constraint on global structure becomes higher\textcolor{black}{.}

\begin{figure*}[ht!]
\begin{tikzpicture}
\pgfplotsset{every tick label/.append style={font=\tiny}}
\begin{axis}[
  title = {\tiny Eiffel\vspace{-1em}},  
  name = eiffel,
  width = 5.5cm,
  height = 5cm,
  enlargelimits=false,   
  xlabel = {},
  ylabel = \empty,
  xtick={0, 2, 4, 6},
  xticklabels= {$N$,  $N-2$,  $N-4$, $N-6$},
  minor y tick num = 1,]
  \addplot[blue!60!black, only marks, mark=+,  mark size=2pt] table[x=x, y=SinGAN] {diversity_measure/eiffel_diversity_comparison_singan.dat};
  \addlegendentry{\tiny SinGAN}
  \addplot[red!60!black, only marks, mark=square, mark size=1pt] table[x=x, y=Ours] {diversity_measure/eiffel_diversity_comparison_singan.dat};
  \addlegendentry{\tiny Ours}
\end{axis}
\begin{axis}[
  title = {\tiny  Musk\vspace{-1em}},  
  name = musk,
  at={($(eiffel.east)+(0.5cm,0)$)},
  anchor=west,
  width = 5.5cm,
  height = 5cm,
  enlargelimits=false,    
  xlabel = {},
  ylabel = \empty,
  xtick={0, 2, 4, 6},
  xticklabels= {$N$,  $N-2$,  $N-4$, $N-6$},
  minor y tick num = 1,]
  \addplot[blue!60!black, only marks, mark=+,  mark size=2pt] table[x=x, y=SinGAN] {diversity_measure/elonmusk_diversity_comparison_singan.dat};
  \addplot[red!60!black, only marks, mark=square, mark size=1pt] table[x=x, y=Ours] {diversity_measure/elonmusk_diversity_comparison_singan.dat};
\end{axis}
\begin{axis}[
  title = {\tiny  Face\vspace{-1em}},  
  name = face,
  at={($(musk.east)+(0.5cm,0)$)},
  anchor=west,
  width = 5.5cm,
  height = 5cm,
  enlargelimits=false,  
  xlabel = {},
  ylabel = \empty,
  xtick={0, 2, 4, 6},
  xticklabels= {$N$,  $N-2$,  $N-4$, $N-6$},
  minor y tick num = 1,]
  \addplot[blue!60!black, only marks, mark=+,  mark size=2pt] table[x=x, y=SinGAN] {diversity_measure/face_diversity_comparison_singan.dat};
  \addplot[red!60!black, only marks, mark=square, mark size=1pt] table[x=x, y=Ours] {diversity_measure/face_diversity_comparison_singan.dat};
\end{axis}
\begin{axis}[
  title = {\tiny  Kid\vspace{-1em}},  
  name = kid,
  at={($(face.east)+(0.5cm,0)$)},
  anchor=west,
  width = 5.5cm,
  height = 5cm,
  enlargelimits=false,    
  xlabel = {},
  ylabel = \empty,
  xtick={0, 2, 4, 6},
  xticklabels= {$N$,  $N-2$,  $N-4$, $N-6$},
  minor y tick num = 1,]
  \addplot[blue!60!black, only marks, mark=+,  mark size=2pt] table[x=x, y=SinGAN] {diversity_measure/kid_diversity_comparison_singan.dat};
  \addplot[red!60!black, only marks, mark=square, mark size=1pt] table[x=x, y=Ours] {diversity_measure/kid_diversity_comparison_singan.dat};
  \vspace{-1.5\baselineskip}
\end{axis}
\end{tikzpicture}
\caption{Diversity (mean std. of pixel values of 50 generated images, higher the better diversity) with scale: ours show more diversity even at lower scales compared to SinGAN~\cite{SHAHAM19}.}
\label{fi:diversity_with_scale}
\end{figure*}
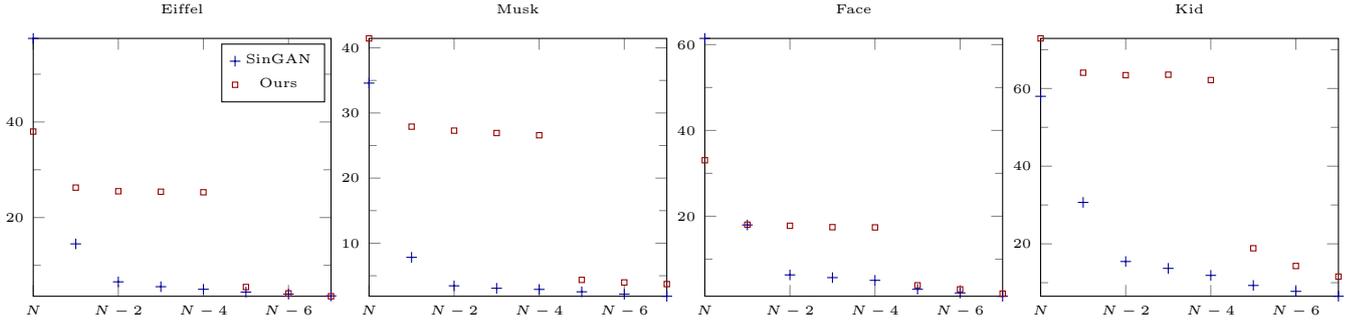
\begin{figure}[t!]
    \centering
    \begin{subfigure}[t]{0.24\linewidth}
        \includegraphics[width=\linewidth]{images/elonmusk1.png}\\
        \vspace{-1.2em}
        \includegraphics[width=\linewidth]{images/kid1.png}\\
        \vspace{-1.2em}
        \includegraphics[width=\linewidth]{images/monaliza1.png}\\
        \vspace{-1.2em}
        \includegraphics[width=\linewidth]{images/eiffel1.png}\\
        \vspace{-1.2em}        
        \includegraphics[width=\linewidth]{images/face1.png}     
        \caption{Input}
    \end{subfigure}%
    \hspace{0.1em}%
    \begin{subfigure}[t]{0.24\linewidth}
        \includegraphics[width=\linewidth]{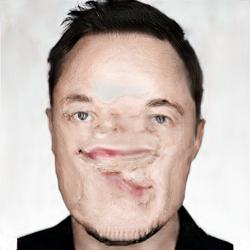}\\
        \vspace{-1.2em}
        \includegraphics[width=\linewidth]{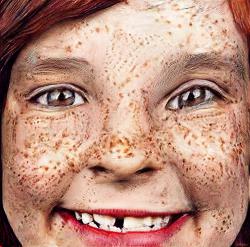}\\
        \vspace{-1.2em}
        \includegraphics[width=\linewidth]{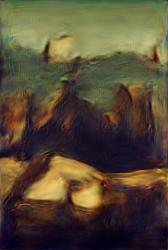}\\
        \vspace{-1.2em}
        \includegraphics[width=\linewidth]{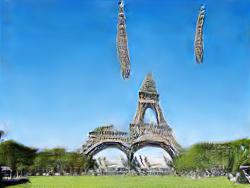}\\
        \vspace{-1.2em}
        \includegraphics[width=\linewidth]{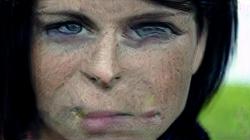}     
        \caption{ConSinGANDef}
    \end{subfigure}%
    \hspace{0.1em}%
    \begin{subfigure}[t]{0.24\linewidth}
        \includegraphics[width=\linewidth]{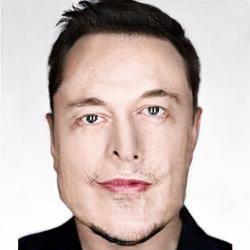}\\
        \vspace{-1.2em}
        \includegraphics[width=\linewidth]{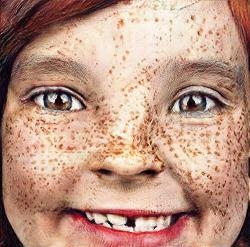}\\
        \vspace{-1.2em}
        \includegraphics[width=\linewidth]{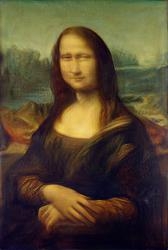}\\
        \vspace{-1.2em}
        \includegraphics[width=\linewidth]{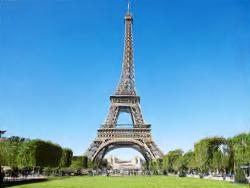}\\
        \vspace{-1.2em}
        \includegraphics[width=\linewidth]{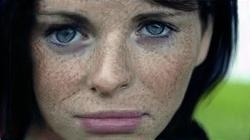}      
        \caption{ConSinGAN6000}
    \end{subfigure}%
    \hspace{0.1em}%
    \begin{subfigure}[t]{0.24\linewidth}
        \includegraphics[width=\linewidth]{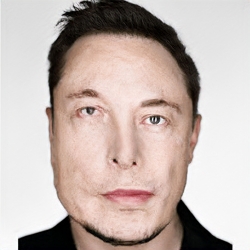}\\
        \vspace{-1.2em}
        \includegraphics[width=\linewidth]{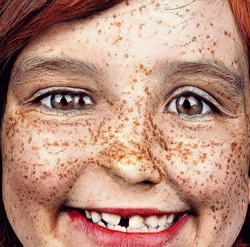}\\
        \vspace{-1.2em}
        \includegraphics[width=\linewidth]{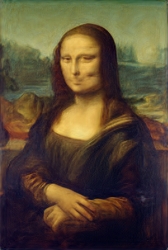}\\
        \vspace{-1.2em}
        \includegraphics[width=\linewidth]{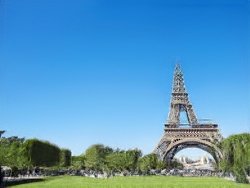}\\
        \vspace{-1.2em}
        \includegraphics[width=\linewidth]{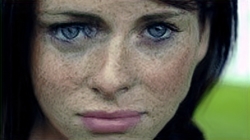}     
        \caption{Ours}
    \end{subfigure} 
    \begin{subfigure}[t]{0.24\linewidth}
        \includegraphics[width=\linewidth]{images/balloons1.png}\\
        \vspace{-1.2em}
        \includegraphics[width=\linewidth]{images/lightning1.png}\\
        \vspace{-1.2em}
        \includegraphics[width=\linewidth]{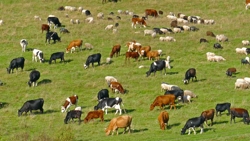}\\
        \vspace{-1.2em}
        \includegraphics[width=\linewidth]{images/colusseum1.png}
        \caption{Input}
    \end{subfigure}%
    \hspace{0.1em}%
    \begin{subfigure}[t]{0.24\linewidth}
        \includegraphics[width=\linewidth]{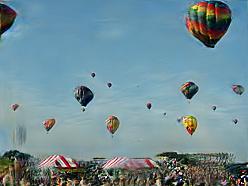}\\
        \vspace{-1.2em}
        \includegraphics[width=\linewidth]{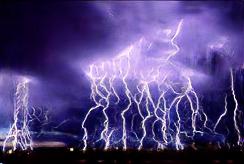}\\
        \vspace{-1.2em}
        \includegraphics[width=\linewidth]{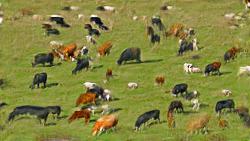}\\
        \vspace{-1.2em}
        \includegraphics[width=\linewidth]{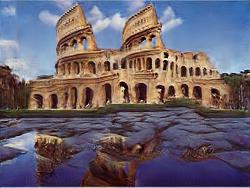}
        \caption{ConSinGANDef}
    \end{subfigure}%
    \hspace{0.1em}%
    \begin{subfigure}[t]{0.24\linewidth}
        \includegraphics[width=\linewidth]{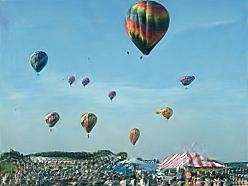}\\
        \vspace{-1.2em}
        \includegraphics[width=\linewidth]{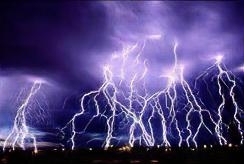}\\
        \vspace{-1.2em}
        \includegraphics[width=\linewidth]{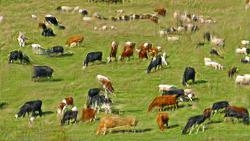}\\
        \vspace{-1.2em}
        \includegraphics[width=\linewidth]{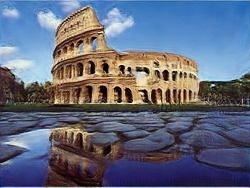}
        \caption{ConSinGAN6000}
    \end{subfigure}%
    \hspace{0.1em}%
    \begin{subfigure}[t]{0.24\linewidth}
        \includegraphics[width=\linewidth]{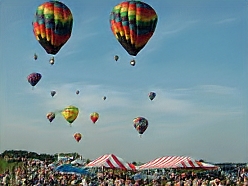}\\
        \vspace{-1.2em}
        \includegraphics[width=\linewidth]{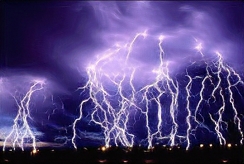}\\
        \vspace{-1.2em}
        \includegraphics[width=\linewidth]{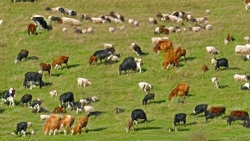}\\
        \vspace{-1.2em}
        \includegraphics[width=\linewidth]{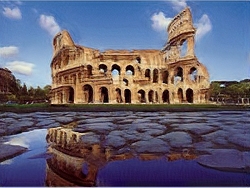}
        \caption{Ours}
    \end{subfigure}%
    \caption{ConSinGAN vs. ours: (b) and (f) show that ConSinGAN with default parameters ( 2000 epochs without any normalization layers inside the architecture ) is not able to preserve the global structure in generating.  Columns (c) and (g) match the hyper-parameters with our with the same number of epochs and instance normalization.}
    \vspace{-0.9\baselineskip}
    \label{fi:consingan_vs_ours}                    
\end{figure}
\begin{table}[t!]
\centering
\caption{Average SIFID score (\textcolor{black}{lower the better}) of generated images from SinGAN~\cite{SHAHAM19} and ours starting with scale $N$ and $N-1$. }
\label{table:1}
\begin{tabular}{@{}l c r@{}}
\toprule
 Generation starting scale & SinGAN & Ours \\ 
 \midrule
 $N$ & $0.02371$ & $0.01828$ \\  
$N-1$ & $0.01396$ & $0.01120$ \\
\bottomrule
\end{tabular}
\vspace{-1em}
\end{table}
 
 We explain how diversity and image quality vary between SinGAN and ours. Diversity is computed among 50 generated images by as the channel wise std. of each pixel and averaging. Fig.~\ref{fi:diversity_with_scale} compares diversity of images which are generated from scale $N$ to scale $N-7$.  We use SA blocks in first 4 scales. These scales are trained with random Gaussian blurred input to maintain diversity compared to the corresponding scales in SinGAN. SinGAN is able to generate images with global context if it starts the generation from scale $N-1$ or above, but it looses the diversity severely compared to the generation from coarsest scale with random noise. So, in SinGAN the maximum achievable diversity with global context can only be attained at scale $N-1$. In all above images the diversity of generated images of SinGAN from scale $N-1$ is lower than the diversity of images with global contexts which are generated by our method from scale $N$ and even in scale $N-1$. This validates that our method is able to produce images with global context without loosing diversity as in SinGAN. 
 We use Single Image Frechet Inception Distance (SIFID) as proposed in SinGAN to compare the quality of generated samples from each image. It uses statistics of features from 2nd layer of the inception network~\cite{7298594}. 
 It compares generated single image with real one by calculating distance between distribution which is created by features before the second pooling layer of Inception Network. 
 It is low when the generated image quality is same as the real image. Table~\ref{table:1} shows the average SIFID for the images which need global structure in column d in Fig.~\ref{fi:resutls_global_and_without}. Our method has lower scores in generations from scale $N$ and $N-1$.
 This shows that our method can generate high quality samples compared to SinGAN in the context of images which need global structure. 
 In summary, the low SIFID and high diversity show that our results are of better quality than SinGAN.

\begin{table}[]
\caption{Comparison of SinGAN~\cite{SHAHAM19} baseline with our proposed improvements in terms of SIFID (lower the better). SinGAN-G: SinGAN with  Gaussian smoothing the input to the discriminator. SinGAN-F: SinGAN with adversarial feedback. 
Adversarial feedback only, and Gaussian smoothing only improve the generation quality in most of the image. 
}
\centering
\label{table:effect_of_individual_components}
\begin{tabular}{@{}lp{1cm}p{1cm}p{1cm}R{1cm}@{}}
\toprule
Image & \multicolumn{1}{l}{SinGAN} & \multicolumn{1}{p{1cm}}{\centering SinGAN \\ + G} & \multicolumn{1}{R{1cm}@{}}{\centering SinGAN \\ + F} \\ 
\midrule
balloons & 0.049  & 0.053 & \textbf{0.041} \\
birds & 0.029 & 0.022 & \textbf{0.021} \\
Colosseum & 0.044  & \textbf{0.038} & 0.038 \\
cows & \textbf{0.037}  & 0.047 & 0.063 \\
Eiffel & 0.041 & \textbf{0.031} & 0.032 \\
Elon Musk & 0.014  & \textbf{0.005} & 0.007 \\
face & 0.031  & 0.018 & \textbf{0.016} \\
kid & \textbf{0.022}  & 0.037 & 0.025 \\
lightning1 & 0.065  & 0.043 & \textbf{0.031} \\
Mona Lisa & 0.03  & 0.047 & \textbf{0.026} \\
mountains & 0.143  & 0.038 & \textbf{0.034} \\
tree & 0.027  & 0.015 & \textbf{0.012} \\
Mean SIFID & 0.044  & 0.033 & \textbf{0.029} \\ 
\bottomrule
\end{tabular}
\end{table}

\begin{table*}[]
\caption{SIFID measured (lower the better). Ours outperforms SinGAN and ConSinGAN  all except in one.}
\label{SIFID_scores}
\centering
\begin{tabular}{@{}L{2cm}R{2cm}R{3cm}R{3cm}R{3cm}R{2cm}@{}}
\toprule
Image & SinGAN &
ConSinGAN Def. Param. &
ConSinGAN 6000 &
ConSinGAN lr\_scale\_0.5 &
 Ours \\
\midrule
balloons & 0.049 & 0.167 & 0.282 & 0.104  & \textbf{0.042} \\
birds & 0.029 & 0.112 & 0.221 & 0.131  & \textbf{0.014} \\
Colosseum & 0.044 & 0.103 & 0.144 & 0.056  & \textbf{0.029} \\
cows & 0.037 & 0.125 & 0.097 & 0.126  & \textbf{0.022} \\
Eiffel & 0.041 & 0.139 & 0.047 & 0.033  & \textbf{0.027} \\
Elon Musk & 0.014 & 0.031 & 0.03 & 0.023  & \textbf{0.004} \\
face & 0.031 & 0.040 & 0.045 & 0.032  & \textbf{0.012} \\
kid & \textbf{0.022} & 0.104 & 0.113 & 0.092 & 0.029 \\
lightning1 & 0.065 & 0.111 & 0.123 & 0.086  & \textbf{0.024} \\
Mona Lisa & 0.030 & 0.139 & 0.099 & 0.141  & \textbf{0.012} \\
mountains & 0.143 & 0.154 & 0.146 & 0.096  & \textbf{0.037} \\
tree & 0.027 & 0.059 & 0.079 & 0.045  & \textbf{0.018} \\
Average & 0.044 & 0.107 & 0.119 & 0.08  & \textbf{0.023}\\
\bottomrule
\end{tabular}
\end{table*}

\begin{table*}[]
\caption{Diversity measured using the average standard deviation of pixel. Note that the diversity in our work  is better when compared with ConSinGAN with the same hyper-parameters needed for maintaining the global structure. Here the diversity values are normalized with mean of pixel values in each image.}
\label{Diversity_scores}
\centering
\begin{tabular}{@{}L{2cm}R{2cm}R{3cm}R{3cm}R{3cm}R{2cm}@{}}
\toprule
Image & SinGAN &
ConSinGAN Def. Param. &
ConSinGAN 6000 &
ConSinGAN lr\_scale\_0.5 &
 Ours \\
 \midrule
balloons & 0.591 & \textbf{0.628} & 0.418 & 0.584 & 0.576 \\
birds & \textbf{0.35} & 0.346 & 0.205 & 0.224 & 0.285 \\
Colosseum & \textbf{0.834} & 0.803 & 0.568 & 0.455 & 0.777 \\
cows & \textbf{0.834} & 0.764 & 0.602 & 0.75 & 0.779 \\
Eiffel & 0.42 & \textbf{0.487} & 0.208 & 0.203 & 0.370 \\
Elon Musk & 0.216 & \textbf{0.254} & 0.192 & 0.169 & 0.201 \\
face & \textbf{0.772} & 0.502 & 0.362 & 0.302  & 0.296 \\
kid & \textbf{0.555} & 0.462 & 0.369 & 0.404  & 0.446 \\
lightning1 & \textbf{1.548} & 1.014  & 0.628 & 0.629 & 0.823 \\
Mona Lisa & \textbf{1.172} & 1.055 & 0.422 & 0.907 & 0.634 \\
mountains & 0.791 & \textbf{0.855} & 0.557 & 0.513 & 0.527 \\
tree & 0.312 & \textbf{0.370} & 0.313 & 0.305 & 0.221 \\
Average & \textbf{0.700} & 0.628 & 0.408 & 0.454 & 0.495\\
\bottomrule
\end{tabular}
\end{table*}

\begin{figure}[t!]
    \centering
    \begin{subfigure}[t]{0.49\linewidth}
        \includegraphics[width=\linewidth]{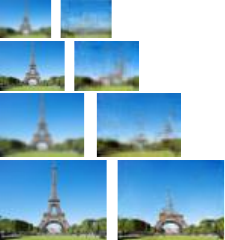}
        \caption{SA at the first three scales ($k=2$),  $\sigma \in [0.5,1]$}
    \end{subfigure}%
    \hspace{0.1em}%
    \begin{subfigure}[t]{0.49\linewidth}
        \includegraphics[width=\linewidth]{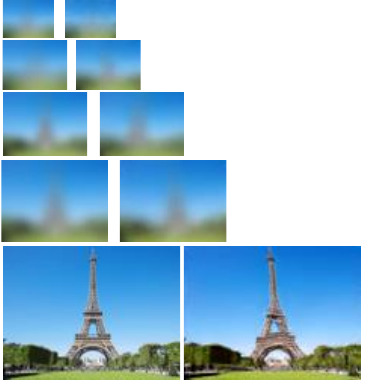}
        \caption{SA at the first four scales ($k=3$),  $\sigma \in [3,7]$}
    \end{subfigure}%
    \caption{Selection of hyper-parameters $k$ and $\sigma$. $k$ is the number of scales having self-attention starting from the coarsest scale $k=0$. Column 1 and 3 are the the Gaussian smoothed real images (smoothed until $N-k$ scale and second and fourth are the corresponding generated results. $\sigma$ is the standard deviation of the smoothing filter applied to the real images given as an input to the discriminator. $k=3$ retains the global structure at $N-4$th scale. Increasing $k$ requires increasing the $\sigma$ of the Gaussian smoothing too to have diversity in image generation.}
    \label{fi:scale}    
\end{figure}
\subsection{Selection of Hyper-Parameters}

Fig.~\ref{fi:scale} show pairs of training and fake images at the end of training at each scale. Column a shows the results of using SA block in first three scales. This $G$ is not able to capture the global structure from the scales with self attention. Here the self attention blocks are added to first three coarser scales [$N,N-1,N-2$]. So, the generations after this scale are not able to preserve the global structure. In the next step, we add SA to scale $N-3$ and to compensate for the reduction of diversity, we increase the standard deviation$\sigma$ range from $[0.5,1]$ to $[3,7]$ . It assists $G$ to capture the global structure fully, which makes the following generations realistic.

\subsection{Impact of Using  Gaussian Kernel with Random Standard Deviation:}

\begin{figure}[t!]
    \centering
    \begin{subfigure}[t]{0.24\linewidth}
        \includegraphics[width=\linewidth]{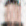}\\
        \vspace{4.8em}
        \caption{scale N}
    \end{subfigure}%
    \hspace{0.1em}%
    \begin{subfigure}[t]{0.24\linewidth}
        \includegraphics[width=\linewidth]{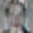}\\
        \vspace{-1.2em}
        \includegraphics[width=\linewidth]{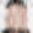}\\
        \vspace{-1.2em}
        \caption{scale N-1}
    \end{subfigure}%
    \hspace{0.1em}%
    \begin{subfigure}[t]{0.24\linewidth}
        \includegraphics[width=\linewidth]{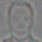}\\
        \vspace{-1.2em}
        \includegraphics[width=\linewidth]{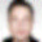}\\
        \vspace{-1.2em}
        \caption{scale N-2}
    \end{subfigure}%
    \hspace{0.1em}%
    \begin{subfigure}[t]{0.24\linewidth}
        \includegraphics[width=\linewidth]{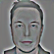}\\
        \vspace{-1.2em}
        \includegraphics[width=\linewidth]{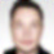}\\
        \vspace{-1.2em}
        \caption{scale N-3}
    \end{subfigure}

    \centering
    \begin{subfigure}[t]{0.24\linewidth}
        \includegraphics[width=\linewidth]{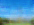}\\
        \vspace{3.6em}
        \caption{scale N}
    \end{subfigure}%
    \hspace{0.1em}%
    \begin{subfigure}[t]{0.24\linewidth}
        \includegraphics[width=\linewidth]{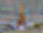}\\
        \vspace{-1.2em}
        \includegraphics[width=\linewidth]{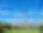}\\
        \vspace{-1.2em}
        \caption{scale N-1}
    \end{subfigure}%
    \hspace{0.1em}%
    \begin{subfigure}[t]{0.24\linewidth}
        \includegraphics[width=\linewidth]{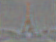}\\
        \vspace{-1.2em}
        \includegraphics[width=\linewidth]{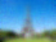}\\
        \vspace{-1.2em}
        \caption{scale N-2}
    \end{subfigure}%
    \hspace{0.1em}%
    \begin{subfigure}[t]{0.24\linewidth}
        \includegraphics[width=\linewidth]{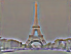}\\
        \vspace{-1.2em}
        \includegraphics[width=\linewidth]{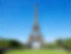}\\
        \vspace{-1.2em}
        \caption{scale N-3}
    \end{subfigure} 
    
    \caption{Intermediate outputs: residual term (Fig.~\ref{fi:overall_architecture} ) from generator at each scale and up-sampled results from previous scale for reconstructed samples.
    Scale $N$ (coarsest) output is an image-like output. However, higher scales generate resisidual (three top right images). Bottom rows show the summation between the residual and up-sampled images form the immediate lower scales.  From scale $N-3$ onward, the network generates the high frequency terms which are subdued in previous scale due to the Gaussian smoothing. As a result the quality of the image generation is not affected by the Gaussian smoothing essential to maintain the diversity.}
    \vspace{-0.9\baselineskip}
    \label{fi:intermediate results}                    
\end{figure}

\begin{figure}[t!]
    \centering
    \includegraphics[bb=0 0 400 200, width=0.98\linewidth]{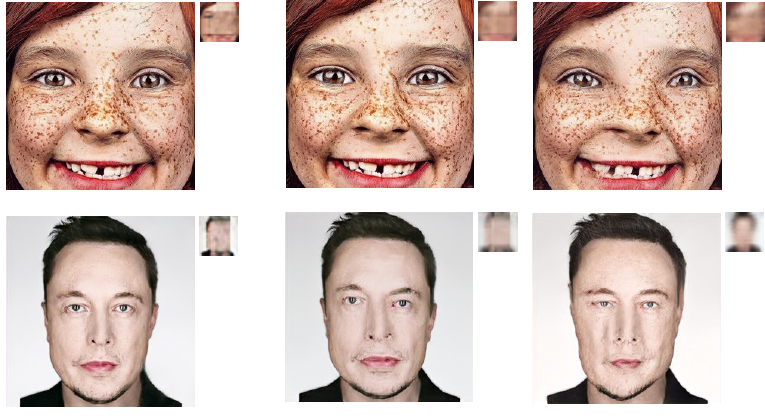}
    \caption{Impact of the $\sigma$ of the Gaussian kernel in generating diversity while maintaining global structure, particularly at the corners.  Column 1 is with no smoothing, column 2 is with $\sigma \in [1,3]$ and column 3 is with $\sigma \in [3,7]$. Form the diversity of hair at top-left corner in column 3,  in comparison with column 1 and 2, we can see that Gaussian kernel with larger sigma encourages diversity particularly near image corners. }    
    \label{fi:impact_of_gauss}   
\end{figure}

\begin{figure}[t!]
    \centering
    \begin{subfigure}[t]{0.33\linewidth}
        \includegraphics[width=\linewidth]{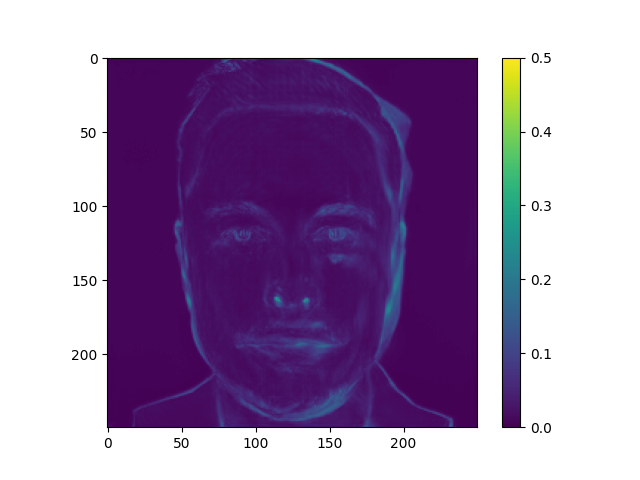}\\
        \vspace{-1.2em}
        \caption{SA only}
    \end{subfigure}%
    \hspace{0.1em}%
    \begin{subfigure}[t]{0.33\linewidth}
        \includegraphics[width=\linewidth]{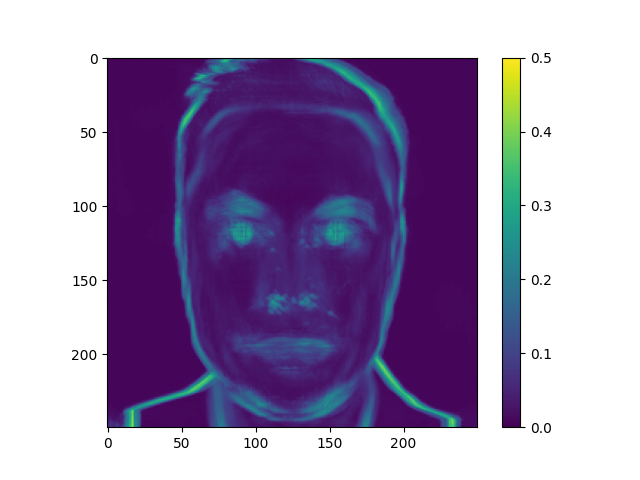}\\
        \vspace{-1.2em}
        \caption{SA3 $\sigma$[1,3]}
    \end{subfigure}%
    \hspace{0.1em}%
    \begin{subfigure}[t]{0.33\linewidth}
        \includegraphics[width=\linewidth]{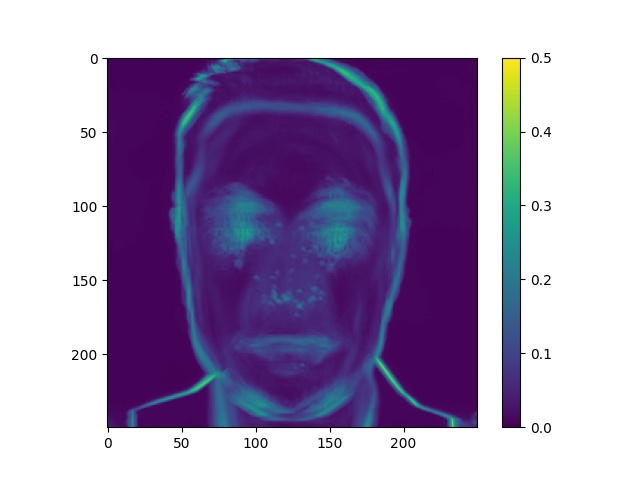}\\
        \vspace{-1.2em}
        \caption{SA3 $\sigma$[3,7]}
    \end{subfigure}%
    \caption{The diversity (as shown by the average std. of pixels among 50 generated images). (a) Only with self-attention. (b) Self-attention and Gaussian smoothing. (c) Self-attention and Gaussian smoothing with a larger $\sigma$.  Notice that as the $\sigma$ of the Gaussian kernel increases, the diversity too increases. }
    \label{fi:spatial_diversity} 
\end{figure}

While using self attention (without random Gaussian blur) the system captures the portion of the image very well from the coarsest scale due to the SA blocks and keeps that portion unchanged in generation. Random Gaussian blur mitigates this issue and adds more diversity. Intermediate results are shown in Fig.~\ref{fi:intermediate results} for two training samples. First row shows the images from the generator output (residual term) and second row shows the upsampled images from previous scale. Here, self attention blocks with random Gaussian augmentation is applied to first 3 scales. In scale 3, the network with a stack of convolution layers with small receptive field is able to learn the remaining high frequency residual term according to its low frequency input from the previous scale.
Fig.~\ref{fi:impact_of_gauss} shows the impact of using the Gaussian smoothed input (see Fig.~\ref{fi:overall_architecture}) in generating images while maintaining the global structure. The figure shows both the sale-0 generation (small images) and the last scale (large images, scale-8). Column 1 is with no smoothing, column 2 is with $\sigma \in [1,3]$ and column 3 is with $\sigma \in [3,7]$. All the three experiments use self-attention in first three scales. Experiments without Gaussian smoothing memorize portion of trained image and have that portion unchanged in  all over the generation (e.g., see the top left corner of column 3 where large $\sigma$ has been successful in generating an image with a large variation in hair in comparison with column 1 and 2). This effect is reduced when adding Gaussian smoothing with larger values of $\sigma$. This effect is clearly shown in Fig.~\ref{fi:spatial_diversity}. Here, the diversity is computed at each spatial location using generated images. Self attention, important to add global information to the generation,  has a disadvantage of reducing the diversity. Gaussian smoothing at the discriminator resolves this issue and increases the diversity. In view of this, providing the Gaussian  is crucial for maintaining diversity in generation. 
The third column (SinGAN + G) of Table~\ref{table:effect_of_individual_components} shows the SIFID scores for the experiments only with Gaussian smoothing. Gaussian smoothing helps to increase the image quality in most of the tested images compared to the SinGAN base line.

\subsection{Impact of Using Adversarial Feedback}
\begin{figure}[t!]
    \centering
    \begin{subfigure}[t]{0.32\linewidth}
        \includegraphics[width=0.5\linewidth]{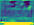}
    \end{subfigure}%
    \hspace{0.1em}%
    \begin{subfigure}[t]{0.32\linewidth}
        \includegraphics[width=0.5\linewidth]{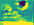}
    \end{subfigure}%
    \hspace{0.1em}%
    \begin{subfigure}[t]{0.32\linewidth}
        \includegraphics[width=0.5\linewidth]{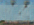}
    \end{subfigure}%
    \newline
    \begin{subfigure}[t]{0.32\linewidth}
        \includegraphics[width=0.75\linewidth]{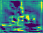}
    \end{subfigure}%
    \hspace{0.1em}%
    \begin{subfigure}[t]{0.32\linewidth}
        \includegraphics[width=0.75\linewidth]{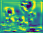}
    \end{subfigure}%
    \hspace{0.1em}%
    \begin{subfigure}[t]{0.32\linewidth}
        \includegraphics[width=0.75\linewidth]{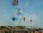}
    \end{subfigure}%
    \newline
    \begin{subfigure}[t]{0.32\linewidth}
        \includegraphics[width=\linewidth]{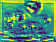}
        \caption{Discriminator score for fake image}
    \end{subfigure}%
    \hspace{0.1em}%
    \begin{subfigure}[t]{0.32\linewidth}
        \includegraphics[width=\linewidth]{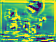}
        \caption{Discriminator score for real image}
    \end{subfigure}%
    \hspace{0.1em}%
    \begin{subfigure}[t]{0.32\linewidth}
        \includegraphics[width=\linewidth]{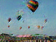}
        \caption{Fake sample}
    \end{subfigure}%
    \caption{Visualization of adversarial feedback of real and fake image at each scale in terms of the discriminator score (blue: low, yellow: high). Top tow: scale 0, mid row: scale 1, bottom row: scale 2. Notice that the regions in the fake image with low quality generation (e.g., balloons) getting lower values in the feedback. Due to the low score, the generator will get a cue where the improvements are needed.}
    \label{fi:effect_of_feedback}    
\end{figure}

\begin{figure}[t!]
    \centering
    \begin{subfigure}[t]{0.49\linewidth}
        \includegraphics[width=\linewidth]{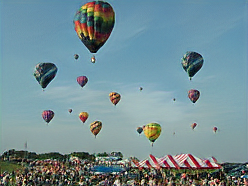}\\
        \vspace{-1.2em}
        \includegraphics[width=\linewidth]{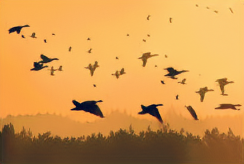}\\
        \vspace{-1.2em}
        \includegraphics[width=\linewidth]{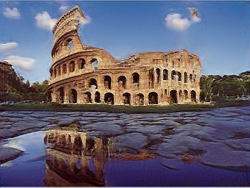}\\
        \vspace{-1.2em}
        \includegraphics[width=\linewidth]{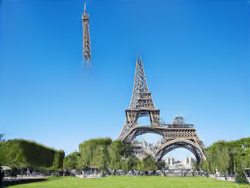}\\
        \vspace{-1.2em}        
        \includegraphics[width=\linewidth]{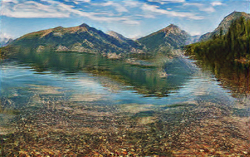}     
        \includegraphics[width=\linewidth]{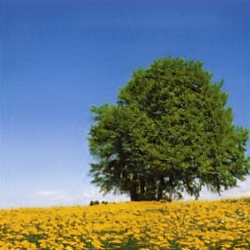}     
        \caption{SinGAN}
    \end{subfigure}%
    \hspace{0.1em}%
    \begin{subfigure}[t]{0.49\linewidth}
        \includegraphics[width=\linewidth]{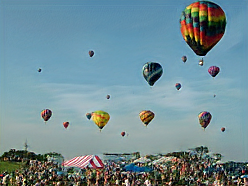}\\
        \vspace{-1.2em}
        \includegraphics[width=\linewidth]{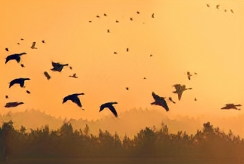}\\
        \vspace{-1.2em}
        \includegraphics[width=\linewidth]{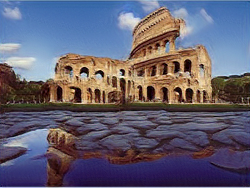}\\
        \vspace{-1.2em}
        \includegraphics[width=\linewidth]{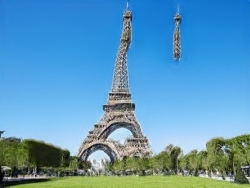}\\
        \vspace{-1.2em}
        \includegraphics[width=\linewidth]{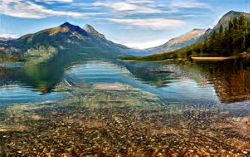}
        \includegraphics[width=\linewidth]{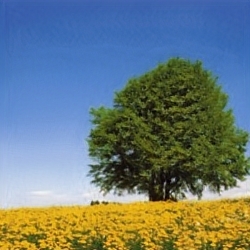}
        \caption{SinGAN with feedback}
    \end{subfigure}%
    \hspace{0.1em}%
    \caption{Effect of incorporating adversarial feedback. The right column with adversarial feedback has better contrast, sharpness (observe the edges) and visual quality. See Table~\ref{table:effect_of_individual_components} for quantitative results.}
    \vspace{-0.9\baselineskip}
    \label{fi:resutls_feedback_GS}                    
\end{figure}

Figure~\ref{fi:effect_of_feedback} shows the visualization of the feedback in terms of discriminator score for three scales of the balloons image. The area with lower quality generation, e.g., balloons in the image, result in low scores in the discriminator. We feed this back to the generator in the next scale (See Fig.~\ref{fi:adversarial_feedback}). Due to the low score, the generator will get a cue where the improvements are needed and vise versa. The last column of Table~\ref{table:effect_of_individual_components} (SinGAN + F) shows the SIFID scores for the experiments only with adversarial feedback. Feedback helps to increase the image quality in most of the tested images. Fig.\ref{fi:resutls_feedback_GS} shows qualitative results of using feedback. 

\subsection{Overall Impact}

\begin{figure*}[t!]
    \centering
    \begin{subfigure}[t]{0.16\linewidth}
        \includegraphics[width=\linewidth]{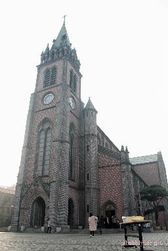}\\
        \vspace{-1.2em}
        \includegraphics[width=\linewidth]{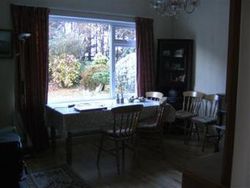}\\
        \vspace{-1.2em}
        \includegraphics[width=\linewidth]{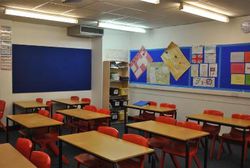}\\
        \vspace{-1.2em}
        \includegraphics[width=\linewidth]{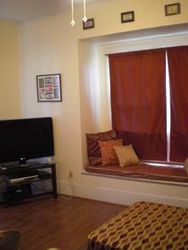}\\
    \end{subfigure}%
    \hspace{0.1em}%
    \begin{subfigure}[t]{0.16\linewidth}
        \includegraphics[width=\linewidth]{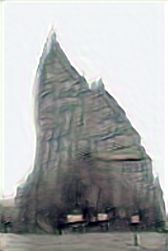}\\
        \vspace{-1.2em}
        \includegraphics[width=\linewidth]{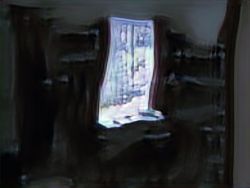}\\
        \vspace{-1.2em}
        \includegraphics[width=\linewidth]{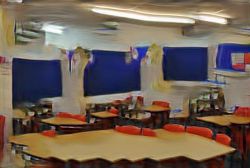}\\
        \vspace{-1.2em}
        \includegraphics[width=\linewidth]{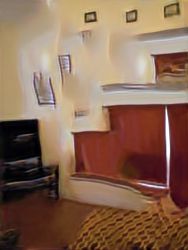}\\
    \end{subfigure}%
    \hspace{0.1em}%
    \begin{subfigure}[t]{0.16\linewidth}
        \includegraphics[width=\linewidth]{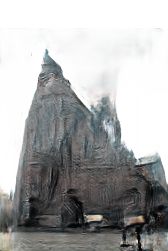}\\
        \vspace{-1.2em}
        \includegraphics[width=\linewidth]{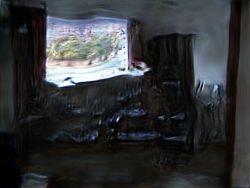}\\
        \vspace{-1.2em}
        \includegraphics[width=\linewidth]{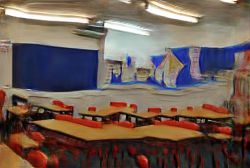}\\
        \vspace{-1.2em}
        \includegraphics[width=\linewidth]{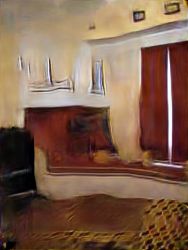}      
    \end{subfigure}%
    \hspace{0.1em}%
    \begin{subfigure}[t]{0.16\linewidth}
        \includegraphics[width=\linewidth]{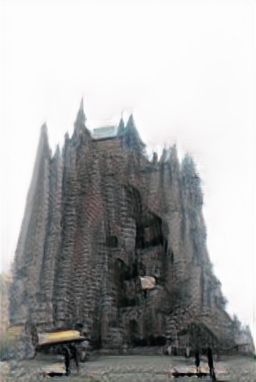}\\
        \vspace{-1.2em}
        \includegraphics[width=\linewidth]{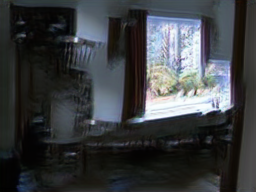}\\
        \vspace{-1.2em}
        \includegraphics[width=\linewidth]{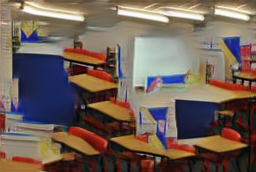}\\
        \vspace{-1.2em}
        \includegraphics[width=\linewidth]{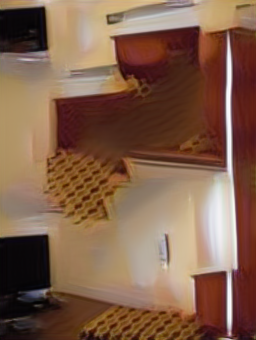}      
    \end{subfigure}%
    \hspace{0.1em}%
    \begin{subfigure}[t]{0.16\linewidth}
        \includegraphics[width=\linewidth]{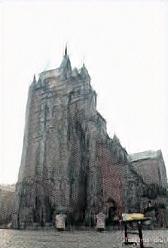}\\
        \vspace{-1.2em}
        \includegraphics[width=\linewidth]{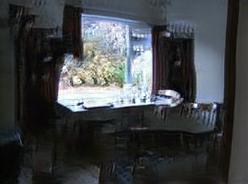}\\
        \vspace{-1.2em}
        \includegraphics[width=\linewidth]{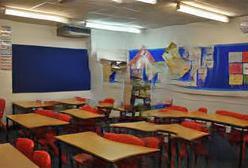}\\
        \vspace{-1.2em}
        \includegraphics[width=\linewidth]{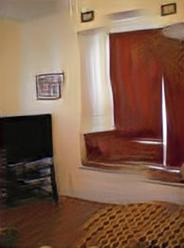}\\
    \end{subfigure} 
    \hspace{0.1em}%
    \begin{subfigure}[t]{0.16\linewidth}
        \includegraphics[width=\linewidth]{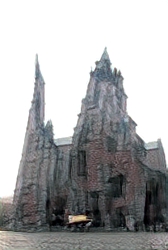}\\
        \vspace{-1.2em}
        \includegraphics[width=\linewidth]{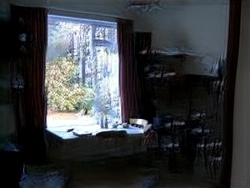}\\
        \vspace{-1.2em}
        \includegraphics[width=\linewidth]{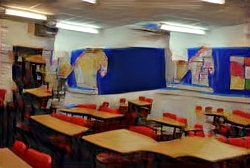}\\
        \vspace{-1.2em}
        \includegraphics[width=\linewidth]{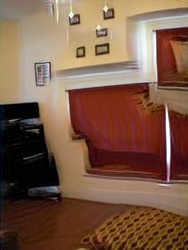}\\
    \end{subfigure} 
       \begin{subfigure}[t]{0.16\linewidth}
        \includegraphics[width=\linewidth]{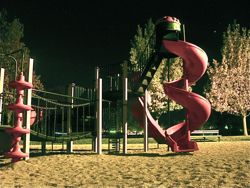}\\
        \vspace{-1.2em}
        \includegraphics[width=\linewidth]{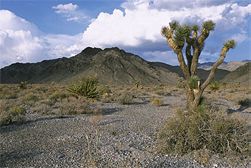}\\
        \vspace{-1.2em}
        \includegraphics[width=\linewidth]{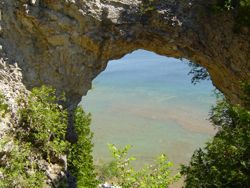}\\
        \vspace{-1.2em}
        \includegraphics[width=\linewidth]{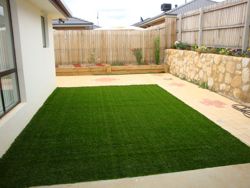}\\
        \caption{Input}
    \end{subfigure}%
    \hspace{0.1em}%
    \begin{subfigure}[t]{0.16\linewidth}
        \includegraphics[width=\linewidth]{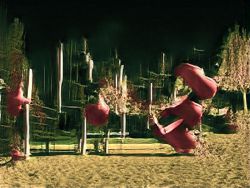}\\
        \vspace{-1.2em}
        \includegraphics[width=\linewidth]{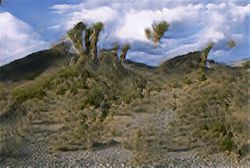}\\
        \vspace{-1.2em}
        \includegraphics[width=\linewidth]{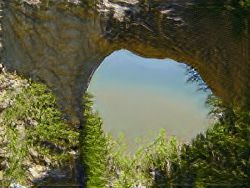}\\
        \vspace{-1.2em}
        \includegraphics[width=\linewidth]{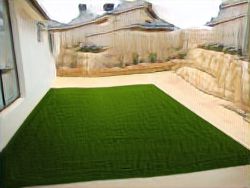}\\
        \caption{SinGAN~\cite{SHAHAM19}}
    \end{subfigure}%
    \hspace{0.1em}%
    \begin{subfigure}[t]{0.16\linewidth}
        \includegraphics[width=\linewidth]{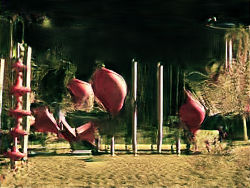}\\
        \vspace{-1.2em}
        \includegraphics[width=\linewidth]{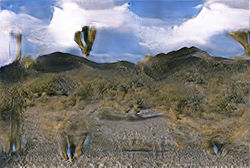}\\
        \vspace{-1.2em}
        \includegraphics[width=\linewidth]{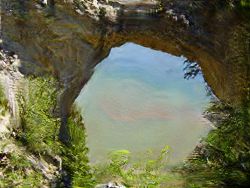}\\
        \vspace{-1.2em}
        \includegraphics[width=\linewidth]{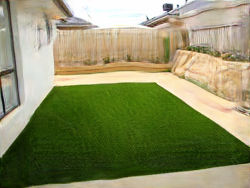}\\     
        \caption{ConSinGAN~\cite{hinz2021improved}}
    \end{subfigure}%
    \hspace{0.1em}%
    \begin{subfigure}[t]{0.16\linewidth}
        \includegraphics[width=\linewidth]{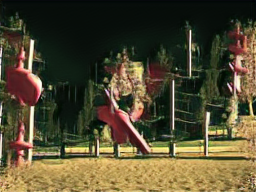}\\
        \vspace{-1.2em}
        \includegraphics[width=\linewidth]{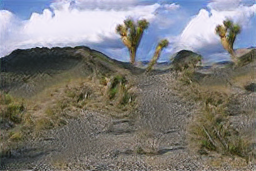}\\
        \vspace{-1.2em}
        \includegraphics[width=\linewidth]{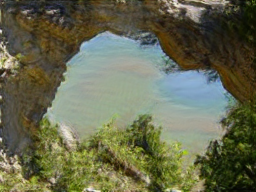}\\
        \vspace{-1.2em}
        \includegraphics[width=\linewidth]{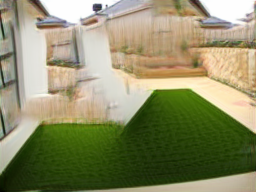}\\      
        \caption{HP-VAE-GAN~\cite{NEURIPS2020_c2f32522}}
    \end{subfigure}%
    \hspace{0.1em}%
    \begin{subfigure}[t]{0.16\linewidth}
        \includegraphics[width=\linewidth]{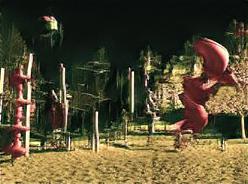}\\
        \vspace{-1.2em}
        \includegraphics[width=\linewidth]{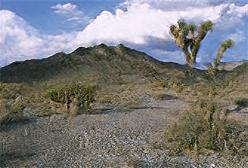}\\
        \vspace{-1.2em}
        \includegraphics[width=\linewidth]{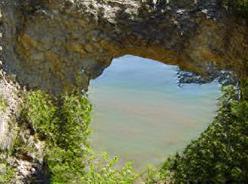}\\
        \vspace{-1.2em}
        \includegraphics[width=\linewidth]{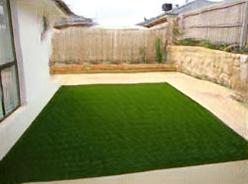}\\     
        \caption{ExSinGAN~\cite{ExSinGAN}}
    \end{subfigure}%
    \hspace{0.1em}%
    \begin{subfigure}[t]{0.16\linewidth}
        \includegraphics[width=\linewidth]{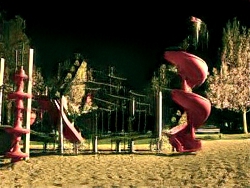}\\
        \vspace{-1.2em}
        \includegraphics[width=\linewidth]{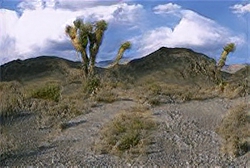}\\
        \vspace{-1.2em}
        \includegraphics[width=\linewidth]{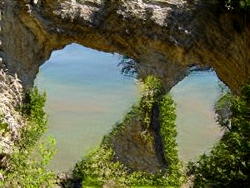}\\
        \vspace{-1.2em}
        \includegraphics[width=\linewidth]{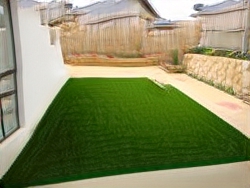}\\
        \caption{Ours}
    \end{subfigure} 
    \caption{\textcolor{black}{Comparision on LSUN~\cite{yu15lsun} and Places~\cite{zhou2017places} datasets. Our method is visually more diverse and \textcolor{black}{achieves} higher quality than the other methods.}}
    \vspace{-0.9\baselineskip}
    \label{fi:lsun_places}                    
\end{figure*}

In this subsection, we compare the performance of our method with SinGAN~\cite{SHAHAM19} and ConSinGAN~\cite{hinz2021improved} with respect to the SIFID scores and the diversity scores. In this experiment we use feedback with self-attention and Gaussian smoothing. With the help of feedback, our method is able to produce realistic diverse results when using self-attention blocks only at coarser scales in first three layers of $G$ and $D$ and Gaussian smoothing with $\sigma$ in the range between $1$ and $3$.

ConSinGAN has been directly developed from SinGAN architecture with the major contribution of training multiple stages concurrently while propagating features within the stages and using a lower learning rate at coarser scale resulted from specific learning rate scale. We first present the average SIFID scores achieved with our method, SinGAN~\cite{SHAHAM19} and ConSinGAN~\cite{hinz2021improved} in Table~\ref{SIFID_scores}. Here, we consider ConSinGAN with the default parameters (2000  epochs with updating $D$ and $G$ three times per epoch, not using normalization layers, using learning rate scale of 0.1), ConSinGAN 6000 which matches hyperparameters of our model (6000 epochs with updating $D$ and $G$ one time per epoch, use instance normalization, using learning rate scale of 0.1) and ConSinGAN lr\_scale\_0.5 (Here learning rate of lower scales are reduce by the factor of 0.5. Hence coarser scales are trained with higher learning rates than above configuration. This helps to maintain more global structure ). It is evident that our method outperforms both SinGAN and ConSinGAN for all the considered images except one.  

Table~\ref{Diversity_scores} shows the  diversity-score comparison with SinGAN and ConSinGAN. Our system gets higher diversity when we consider the ConSinGAN 6000 and ConSinGAN lr\_scale\_0.5 which generates realistic samples for the images which need global structure to be maintained. Diversity of our system is lower than ConSinGAN with default parameters, but ConSinGAN generated images are not realistic. When compared to SinGAN, our method looses diversity for constraining in the global structure through self attention blocks. As we mentioned in Fig.~\ref{fi:diversity_with_scale}, our method generates highly diverse images compared to SinGAN when it starts the generation not from the coarser scales to retain global structure.
Fig.~\ref{fi:consingan_vs_ours} shows the visual results comparison with ConSinGAN. ConSinGAN with default parameters is clearly unable to preserve the global structure. Our method constrains the global structure for the images with larger objects like faces, humans, and buildings while generating diverse images with smaller objects like balloons and cows.

\begin{table}[t!]
\centering
\caption{\textcolor{black}{Average SIFID score (lower the better), diversity (higher the better), and user votes (higher the better) of generated images from LSUN dataset}}
\label{table:lsun}
\begin{tabular}{@{}l c c  r@{}}
\toprule
Methods & SIFID $\downarrow$ & Diversity $\uparrow$ & \textcolor{black}{User votes $\uparrow$} \\ 
 \midrule
SinGAN~\cite{SHAHAM19} & $0.11$ & $0.60$ & \textcolor{black}{$0$} \\  
ConSinGAN~\cite{hinz2021improved} & $0.08$ & $0.55$ & \textcolor{black}{$5$} \\
HP-VAE-GAN~\cite{NEURIPS2020_c2f32522} & $0.40$ & $0.78$ & \textcolor{black}{$0$} \\
ExSinGAN~\cite{ExSinGAN} & $0.11$ & $0.50$ & \textcolor{black}{$16$} \\
Ours & $0.06$ & $0.60$ & \textcolor{black}{$28$}\\
\bottomrule
\end{tabular}
\vspace{-1em}
\end{table}

\begin{table}[t!]
\centering
\caption{\textcolor{black}{Average SIFID score (lower the better), diversity (higher the better), and user votes (higher the better) of generated images from Places dataset}}
\label{table:places}
\begin{tabular}{@{}l c c r@{}}
\toprule
Methods & SIFID $\downarrow$ & Diversity $\uparrow$ & \textcolor{black}{User votes $\uparrow$} \\ 
 \midrule
SinGAN~\cite{SHAHAM19} & $0.09$ & $0.52$ & \textcolor{black}{$2$}  \\  
ConSinGAN~\cite{hinz2021improved} & $0.06$ & $0.50$ & \textcolor{black}{$7$}  \\
HP-VAE-GAN~\cite{NEURIPS2020_c2f32522} & $0.17$ & $0.62$ & \textcolor{black}{$1$} \\
ExSinGAN~\cite{ExSinGAN} & $0.10$ & $0.47$ & \textcolor{black}{$16$} \\
Ours & $0.04$ & $0.44$ & \textcolor{black}{$24$} \\
\bottomrule
\end{tabular}
\vspace{-1em}
\end{table}

\textcolor{black}{We also test our method in the same 50 images of LSUN~\cite{yu15lsun} and Places~\cite{zhou2017places} datasets as in the prior works. In Table~\ref{table:lsun} and Table~\ref{table:places} we present the average SIFID and the diversity scores for LSUN and Places datasets, \textcolor{black}{respectively}. In both datasets we achieve lower SIFID scores without degrading much in the diversity. We show the qualitative comparison in Figure~\ref{fi:lsun_places}.} 

\textcolor{black}{Furthermore, we conducted a user study to evaluate the quality and diversity of generated images. Using Amazon Mechanical Turk (AMT), we showed the original image and 5 generated images from different methods (SinGAN\cite{SHAHAM19}, ConSinGAN\cite{hinz2021improved}, ExSinGAN\cite{ExSinGAN}, HP-VAE-GAN\cite{NEURIPS2020_c2f32522} and our method) and requested the users to select the one with the highest quality and diversity. We showed 49 images from LSUN~\cite{yu15lsun} and 50 images of Places~\cite{zhou2017places} for this user study, and we used one image from LSUN as an example for the users to explain the task clearly. We used AMT to aggregate the responses from 49 participants. For each image, we assigned a score of 1 for the method having the highest user votes, and we present the overall results in Tables~\ref{table:lsun} and~\ref{table:places}, respectively, for LSUN and Places datasets. For both datasets, our method achieves the highest scores confirming that our method generates images with higher quality and diversity compared to previously proposed methods.}






\subsection{Image Harmonization, Editing, and Generating Arbitrary-Sized Images}

\begin{figure}[t!]
    \centering
    \begin{subfigure}[t]{0.49\linewidth}
        \includegraphics[width=\linewidth]{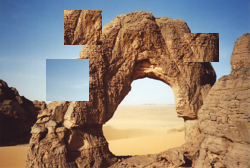}
        \caption{Edited input}
        \label{sfi:editing_input}        
    \end{subfigure}%
    \hspace{0.1em}%
        \begin{subfigure}[t]{0.49\linewidth}
        \includegraphics[width=\linewidth]{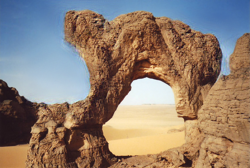}
        \caption{Output due to the edited image}
        \label{sfi:editing_output2}         
    \end{subfigure}%
    \hspace{0.1em}%
        \begin{subfigure}[t]{0.49\linewidth}
        \includegraphics[width=\linewidth]{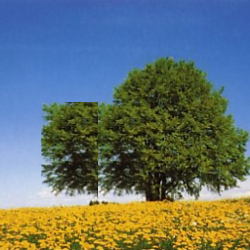}
        \caption{Edited input}
        \label{sfi:editing_output3}         
    \end{subfigure}%
    \hspace{0.1em}%
        \begin{subfigure}[t]{0.49\linewidth}
        \includegraphics[width=\linewidth]{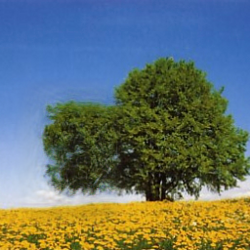}
        \caption{Output due to the edited image}
        \label{sfi:editing_output4}         
    \end{subfigure}%
    
    \caption{The performance of our system in image editing: (a) Edited image: Note the rectangular edit (b) successful removal of the edit}
    \label{fi:image_editing}
\end{figure}    
    
\begin{figure}[t!]
    \begin{subfigure}[t]{0.49\linewidth}
        \includegraphics[width=\linewidth]{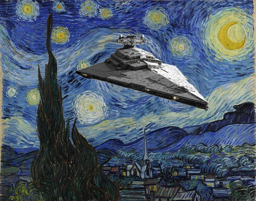}
        \label{sfi:harmonization_input1}
    \end{subfigure}%
    \hspace{0.1em}%
        \begin{subfigure}[t]{0.49\linewidth}
        \includegraphics[width=\linewidth]{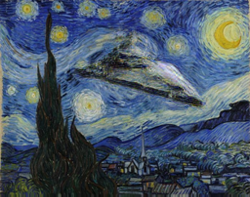}
        \label{sfi:harmonization_output1}
    \end{subfigure}%
    \hspace{0.1em}%
    \begin{subfigure}[t]{0.49\linewidth}
        \includegraphics[width=\linewidth]{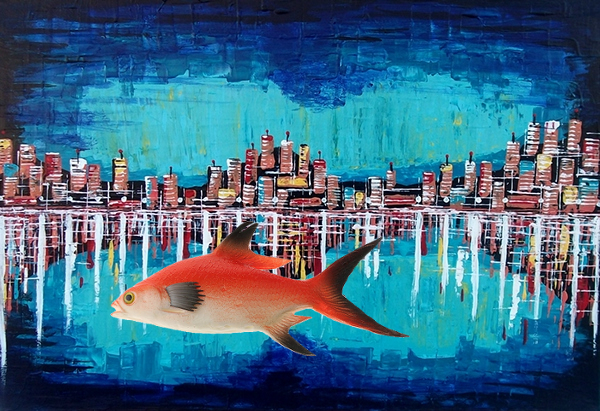}
        \label{sfi:harmonization_input2}
    \end{subfigure}%
    \hspace{0.1em}%
        \begin{subfigure}[t]{0.49\linewidth}
        \includegraphics[width=\linewidth]{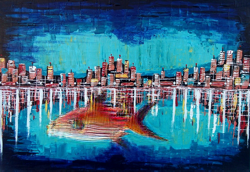}
        \label{sfi:harmonization_output2}
    \end{subfigure}%
    \hspace{0.1em}%
    \begin{subfigure}[t]{0.49\linewidth}
        \includegraphics[width=\linewidth]{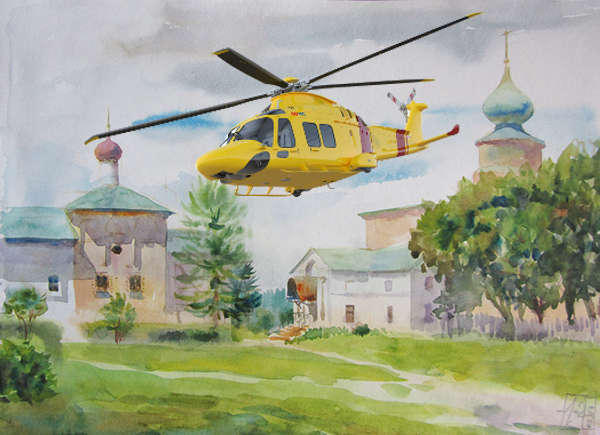}
        \caption{Harmonization input}
        \label{sfi:harmonization_input3}
    \end{subfigure}%
    \hspace{0.1em}%
        \begin{subfigure}[t]{0.49\linewidth}
        \includegraphics[width=\linewidth]{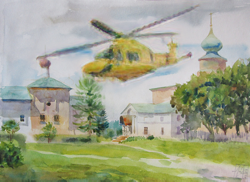}
        \caption{Harmonized output}
        \label{sfi:harmonization_output3}
    \end{subfigure}%
    \caption{The performance of our system in harmonization: New mask from different patch (e.g., spacecraft) statistics is harmonized into the trained image.}
    \label{fi:harmonization}
\end{figure}    
    
\begin{figure}[t!]
    \begin{subfigure}[t]{0.99\linewidth}
        \includegraphics[width=\linewidth]{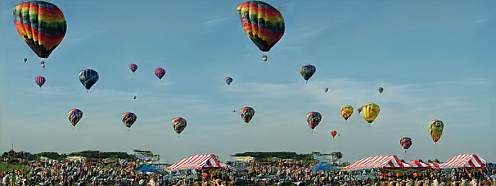}
        \label{sfi:arbitrary_size_generation_1}        
    \end{subfigure}%
    \hspace{0.1em}%
        \begin{subfigure}[t]{0.99\linewidth}
        \includegraphics[width=\linewidth]{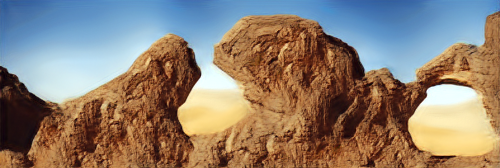}
        \label{sfi:arbitrary_size_generation_2}      
    \end{subfigure}%
     \hspace{0.1em}%
    \begin{subfigure}[t]{0.99\linewidth}
        \includegraphics[width=\linewidth]{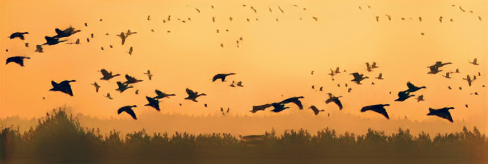}
        \label{sfi:arbitrary_size_generation_3}        
    \end{subfigure}%
     \hspace{0.1em}%
    \begin{subfigure}[t]{0.99\linewidth}
        \includegraphics[width=\linewidth]{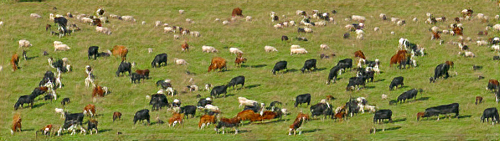}
        \label{sfi:arbitrary_size_generation_4}        
    \end{subfigure}%
    \caption{ Arbitrary sized image generation: The new images are generated from noise with a different aspect ratio from the trained image. In this particular example, the width is doubled. Notice how larger images are generated while preserving the global structure.}
    \label{fi:arbitrary_size_generation}
\end{figure}

Here we show the results of some other tasks with using self-attention blocks, Gaussian smoothing and feedback.  We can do image harmonization by feeding the source image and the image to be blended in at a intermediate scale (e.g., scale 4 or 5). Then the generated image will have the image to be blended harmonized into the source image. Fig.~\ref{fi:harmonization} shows the ability of our system to harmonize images. Notice how, e.g., the space craft has been harmonized into the background image.

In image editing, an artificially inserted patch at a course scale will be blended in without artifacts. Binary mask indicating the location of inserted patch helps to refine the results by only changing the portion of the inserted patch Fig.~\ref{fi:image_editing} shows the ability our our system to edit images. Notice how the edits (light blue patch, and the green branches) have been successfully blended-in.


Our method able to produce arbitrary-sized images, this is because of using fixed scale size ($m$) after downsampling operation in self attention blocks. If we do not use a fixed scale for downsampling, additional features from the arbitray size noise will interfere self attention blocks and reduce the quality of the output. Using a constant scale down size helps to find the inter dependencies of downsampled feature space to compute the re-weighted feature samples. Fig.~\ref{fi:arbitrary_size_generation} show four example of arbitrary-sized image generations. Our method cannot produce realistic results when generating arbitrary size images from faces because network cannot preserve the global structure for the arbitrary size input which is not available at the training time. We observed that our animation results also are better in quality than SinGAN.

Above results show that our method is able to produce results on par with SinGAN. Our attention module does not interfere with the the ability to the system in image harmonization, editing, and generating arbitrary-sized images. We are able to do these tasks while preserving the global structure. 

\begin{figure}[t!]
    \begin{subfigure}[t]{0.21\linewidth}
        \includegraphics[width=\linewidth]{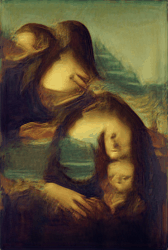}
        \caption{SinGAN~\cite{SHAHAM19}}
    \end{subfigure}%
    \hspace{0.1em}%
    \begin{subfigure}[t]{0.21\linewidth}
        \includegraphics[width=\linewidth]{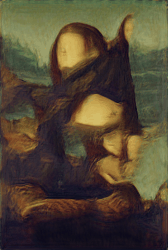}
    \end{subfigure}%
    \hspace{0.1em}%
    \begin{subfigure}[t]{0.21\linewidth}
        \includegraphics[width=\linewidth]{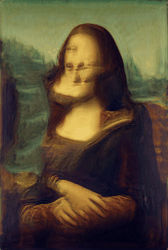}
    \end{subfigure}%
    \hspace{0.1em}%
    \begin{subfigure}[t]{0.21\linewidth}
        \includegraphics[width=\linewidth]{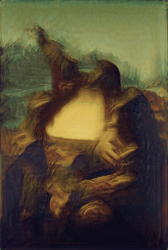}
    \end{subfigure}%
    \hspace{0.1em}%
    \begin{subfigure}[t]{0.21\linewidth}
        \includegraphics[width=\linewidth]{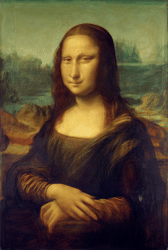}
        \caption{Ours}
    \end{subfigure}%
    \hspace{0.1em}%
    \begin{subfigure}[t]{0.21\linewidth}
        \includegraphics[width=\linewidth]{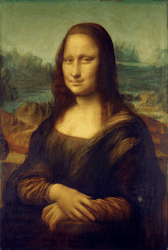}
    \end{subfigure}%
    \hspace{0.1em}%
    \begin{subfigure}[t]{0.21\linewidth}
        \includegraphics[width=\linewidth]{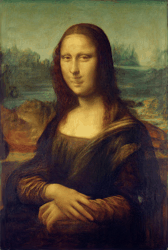}
    \end{subfigure}%
    \hspace{0.1em}%
    \begin{subfigure}[t]{0.21\linewidth}
        \includegraphics[width=\linewidth]{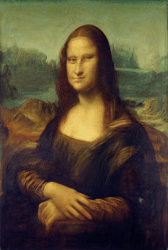}
    \end{subfigure}%
    
    \caption{\textcolor{black}{Four frames of animated videos generated using our method and SinGAN. Our method can generate realistic animations with higher diversity while maintaining the global structure.}}
    \label{fi:Anim}
\end{figure}

\textcolor{black}{\subsection{Animation and Image Augmentation}
Here we consider two applications of the proposed model: animation and image augmentation. We present four frames of a video generated by our method and SinGAN~\cite{SHAHAM19} in Fig.~\ref{fi:Anim}. We clearly observe that our method improves the fidelity of an animated video generated with an image which needs global structure to look realistic. This confirms that our method can generate realistic animations with higher diversity while maintaining the global structure compared to SinGAN.}

\textcolor{black}{We then conduct an experiment to evaluate the feasibility of using our method for data augmentation. To this end, we select two classes (abbey and arch) from SUN database~\cite{5539970}. In each class, we use 4 images and train them to generate 500 images from SinGAN~\cite{SHAHAM19} and our method. Next, we train separate classifiers (ResNet18\cite{7780459}) using these generated images and evaluate the performance with the test set. The classifier trained using the images generated from our method achieves an accuracy of 63.68$\%$ whereas the classifier trained using the images generated from SinGAN achieves an accuracy of only 58.42$\%$. The classifier trained only with 8 original samples achieves 52.11$\%$. These experimental results—although produced with eight images—confirm that our model can be successfully used for image augmentation.}

\section{\textcolor{black}{Limitations}}
\textcolor{black}{Our work augments the SinGAN architecture to improve its generation quality with an additional attention layer and feedback through the discriminator. Compared to SinGAN, attention layers introduce additional number of trainable parameters. Unlike in SinGAN, we need to forward pass through the previous discriminators for generating the feedback. Having the Gaussian smoothing augmentation also adds an overhead. These reasons make our methodology to take higher training time than SinGAN and ConSinGAN. Even though with our default parameters we achieve the better generation quality without degrading the diversity, image specific parameters will lead to more better results. Since our method trains with only one sample, we cannot control the specific semantic characteristics in the generation from noise.}

\section{Conclusion}
\label{sec:foot}
In this work, we improved the technique of image generation in realistic diverse single image generation when global structure is more important. We were able to control the level of global contextual information insertion using self-attention blocks, impose the diversity through convolving the input with a random Gaussian kernel when training the discriminator, and improve the quality of generation with adversarial feedback. This also helped to increase the diversity in generating  samples from less-coarse scales significantly compared to SinGAN. Our future work will address generating images that need the global context with varying aspect ratios, denoising, and image inpainting using this work.

\section*{Acknowledgement}
This work was supported in part by the National Resource Council of Sri Lanka under the grant 19-080.
\bibliography{mybibfile}
\end{document}